\documentclass{article}
\usepackage{ijcai25}

\usepackage{times}
\usepackage{soul}
\usepackage{url}
\usepackage[hidelinks]{hyperref}
\usepackage[utf8]{inputenc}
\usepackage[small]{caption}
\usepackage{graphicx}
\usepackage{amsmath}
\usepackage{amsthm}
\usepackage{booktabs}
\usepackage{algorithm}
\usepackage{algorithmic}
\usepackage[switch]{lineno}
\usepackage{bm}
\usepackage{amssymb}
\usepackage{multirow}
\usepackage{threeparttable}
\usepackage{array}
\usepackage{makecell}

\usepackage{subcaption}
\usepackage{array}

\title{MMET: A Multi-Input and Multi-Scale Transformer for Efficient PDEs Solving}

\author{
	Yichen Luo$^1$\and
	Jia Wang$^2$\and
    Dapeng Lan$^3$\and
	Yu Liu$^3$\And
	Zhibo Pang$^{*,1,4}$\\
	\affiliations
	$^1$Division of ISE, KTH Royal Institute of Technology, Stockholm, Sweden\\
	$^2$School of Advanced Technology, Xi'an Jiaotong-Liverpool University, Suzhou, China\\
	$^3$Techforgood AS, Oslo, Norway\\
	$^4$Department of Automation Technology, ABB Corporate Research Sweden, Vasteras, Sweden\\
	\emails
	\{yichenlu, zhibo\}@kth.se,
	Jia.Wang02@xjtlu.edu.cn,
	\{dapengl, liu.yu\}@ieee.org
}

\begin{document}

\maketitle

\begin{abstract}
    Partial Differential Equations (PDEs) are fundamental for modeling physical systems, yet solving them in a generic and efficient manner using machine learning-based approaches remains challenging due to limited multi-input and multi-scale generalization capabilities, as well as high computational costs. This paper proposes the Multi-input and Multi-scale Efficient Transformer (MMET), a novel framework designed to address the above challenges. MMET decouples mesh and query points as two sequences and feeds them into the encoder and decoder, respectively, and uses a Gated Condition Embedding (GCE) layer to embed input variables or functions with varying dimensions, enabling effective solutions for multi-scale and multi-input problems. Additionally, a Hilbert curve-based reserialization and patch embedding mechanism decrease the input length. This significantly reduces the computational cost when dealing with large-scale geometric models. These innovations enable efficient representations and support multi-scale resolution queries for large-scale and multi-input PDE problems. Experimental evaluations on diverse benchmarks spanning different physical fields demonstrate that MMET outperforms SOTA methods in both accuracy and computational efficiency. This work highlights the potential of MMET as a robust and scalable solution for real-time PDE solving in engineering and physics-based applications, paving the way for future explorations into pre-trained large-scale models in specific domains. This work is open-sourced at \url{https://github.com/YichenLuo-0/MMET}.
\end{abstract}

\section{Introduction}

\begin{figure}[thpb]
    \centering
    \includegraphics[width=\linewidth, trim=25 20 15 15, clip]{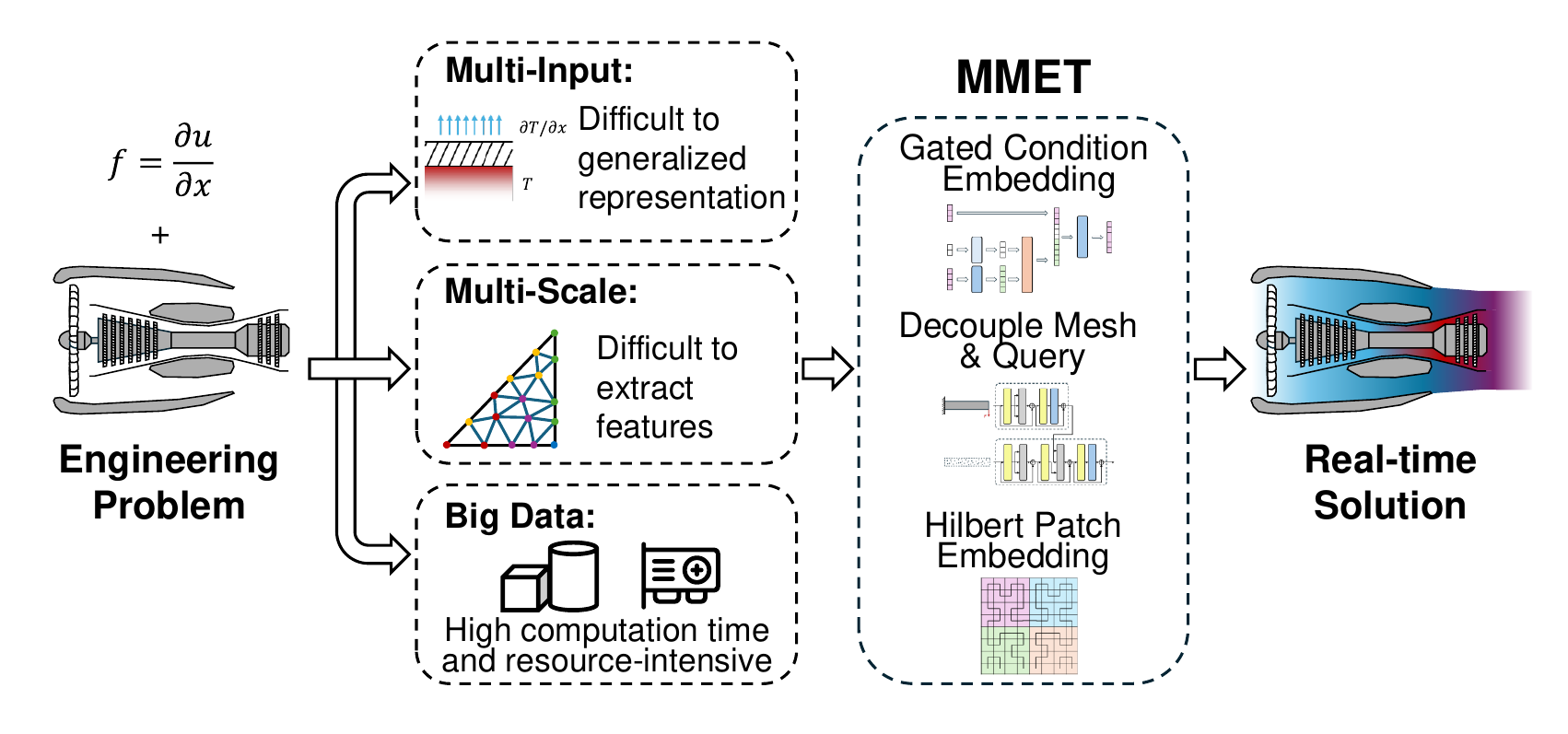}
    \caption{MMET mainly addresses three challenges in applying machine learning to large-scale PDEs solving, including multiple initial/boundary conditions input, multi-scale resolution, and high computational cost in large-scale geometric models.}
    \label{result33}
\end{figure}

The solution of complex Partial Differential Equations (PDEs) has long been one of the great challenges in engineering. In the vast majority of cases involve elasticity, fluid dynamics or other physics fields, it is difficult to obtain analytical solutions for their governing PDEs. As an alternative, numerical methods such as the Finite Element Method (FEM) are usually used to approximate the result. However, these methods are computationally expensive and challenging to accelerate using parallel hardware.

Recent advances in the field of Artificial Intelligence (AI) have explored the use of machine learning techniques to approximate physical laws directly from PDEs \cite{RAISSI2019686} or big data \cite{wang2021learning}, and have demonstrated promising accuracy in Computational Fluid Dynamics (CFD) \cite{Maziar2020Hidden,MAO2020112789}, magnetohydrodynamics \cite{Rosofsky2023Magnet}, and other fields. Nonetheless, traditional fully connected neural networks can typically solve only a single instance in a set of PDEs. Any changes in geometric shape, initial/boundary conditions, or material properties require retraining of the neural network. This makes the speed advantage of AI-based approach over traditional FEM a pseudo proposition, as the cost of training a neural network for a single instance can be comparable to or even exceed that of FEM.

Recently, generative pre-trained large-scale models based on the Transformer architecture \cite{vaswani2023attentionneed} have shown strong zero-shot or few-shot generalization abilities in natural language processing \cite{openai2024gpt4}, machine vision \cite{liu2024sora}, and more scenarios. By extending this paradigm to the field of PDE-driven physics simulation, it becomes feasible to develop universal models that do not require expensive retraining for each instance. Several research efforts have attempted to fit PDEs using the Transformer architecture, targeting various aspects such as enhancing operator learning \cite{Li2022TransformerFP,SHIH2025117560}, generalization \cite{santos2023physicsinformed}, multi-scale meshes \cite{hao2023gnot}, simulation accuracy \cite{zhao2024pinnsformer}, geometric model segmentation \cite{wu2024Transolver}, and improving the attention mechanism to enhance accuracy \cite{cao2021choosetransformerfouriergalerkin}.

Despite these advancements, the above methods primarily address specific challenges within their respective domains. However, a universal paradigm that can systematically tackle all the key challenges within a unified framework remains absent. These challenges can be summarized as follows.

\begin{itemize}
	\item\textbf{Multi-Input Representation:} A complete PDE solving problem can contain multiple inputs, such as differential operator, geometry domain, initial/boundary conditions, etc., which are in the form of variables or functions. The model must possess a robust and adaptable input encoding mechanism capable of efficiently encoding above diverse inputs.

	\item\textbf{Multi-Scale Resolution:} Different scenarios often require varying output resolutions, leading to inconsistencies in query inputs. However, Transformer-based models may alter their attention mechanisms as the input sequence changes, resulting in fluctuations in query accuracy under different resolutions.

	\item\textbf{Computationally Expensive:} Meshes in industrial scenarios often consist of tens to hundreds of thousands of node points, posing significant challenges to the cost and efficiency of attention mechanisms at such scales.
\end{itemize}

Based on these limitations, we propose the \textbf{M}ulti-input and \textbf{M}ulti-scale \textbf{E}fficient \textbf{T}ransformer (\textbf{MMET}) architecture, which can be used to solve a class of general problems in real time for certain groups of PDEs. Our contributions can be summarized as follows:

\begin{itemize}
	\item Proposing the Gated Condition Embedding (GCE) layer, embedding input functions and variables in units of spatio-temporal points, which can effectively encode complex input conditions while ensuring that the input sequence length does not increase significantly.
	
	\item Decoupling the query points and the geometry mesh into two sequences for the input of the encoder and decoder, respectively, enabling queries at any resolution.
	
	\item Reserializing and patching the mesh sequence using the Hilbert curve \cite{hilbert1935stetige}, significantly reducing the computational cost of the attention mechanism.
\end{itemize}

Experiments across various physical fields demonstrate that these innovative methods significantly enhance both the accuracy and computational efficiency of models in solving multi-input and multi-scale PDEs.

\section{Related Work}

\subsection{Physics-Informed Neural Networks}

Physics-Informed Neural Networks (PINNs) \cite{RAISSI2019686} are a class of numerical simulation approaches based on machine learning aiming to approximate the solution of PDEs by incorporating physical laws directly into the loss function of the artificial neural network. This methodology can actually be regarded as a form of self-supervised learning rather than a network design, as the architecture of the neural network itself is highly flexible.

\begin{figure*}[thpb]
	\centering
	\includegraphics[width=\textwidth, trim=10 25 10 25, clip]{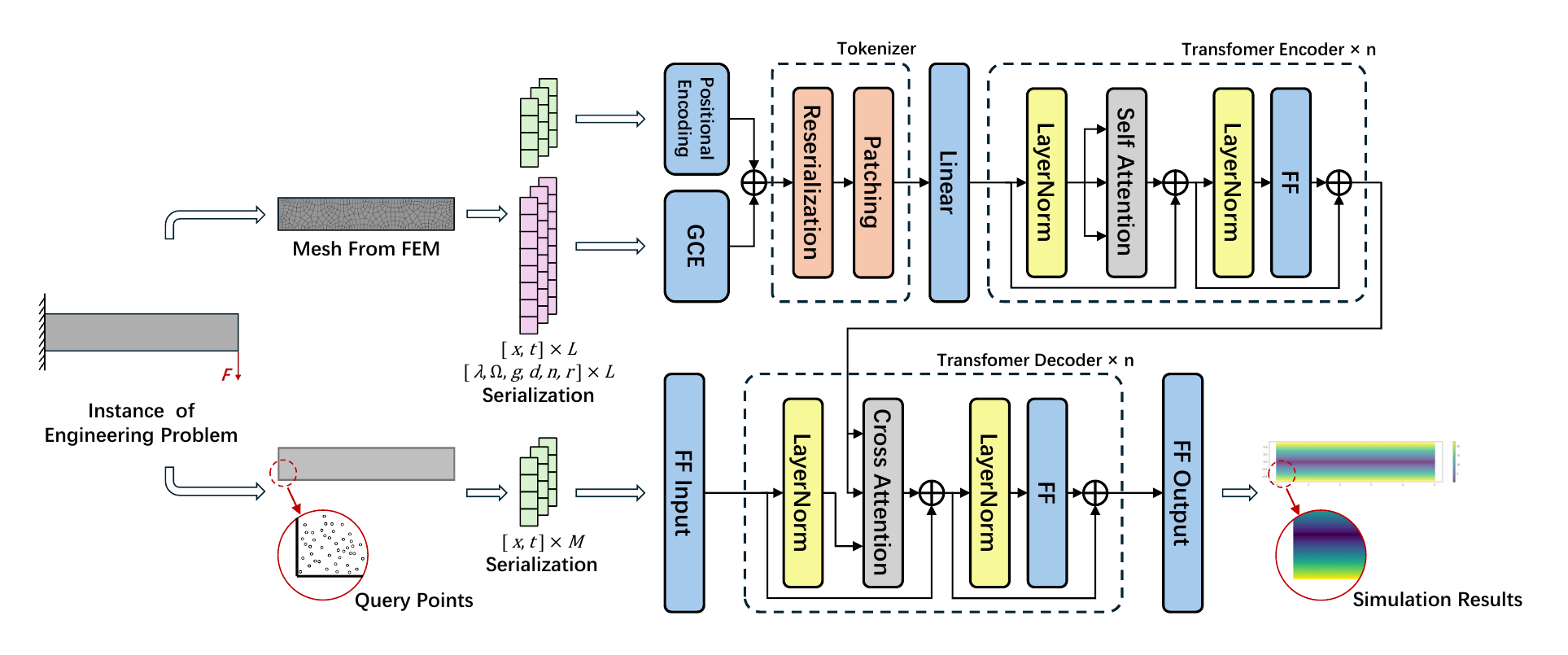}
	\caption{Overall architecture of MMET. Where $L$ and $M$ are the sequence lengths of the serialized mesh and query points, respectively. In this architecture, the encoder is responsible for extracting and encoding the feature information of the whole problem domain, which comes from the mesh pseudo-sequence; and the decoder is responsible for solving the value at the query points based on the encoded information.}
	\label{arch1}
\end{figure*}

\subsection{Operator Learning}

The operator learning paradigm, originating from DeepONet \cite{wang2021learning}, has recently replaced traditional neural network architectures as a new generalized PDE solver. Among these methods, the FNO \cite{Li2020FourierNO} stands out for its efficiency, leveraging linear projections in the Fourier domain to handle high-dimensional inputs. Extensions such as U-NO \cite{rahman2023unoushapedneuraloperators} and F-FNO \cite{tran2023factorizedfourierneuraloperators} improve multiscale representation or computational efficiency, further expanding the versatility of neural operators. On the other side, models like the GNO \cite{li2020neuraloperatorgraphkernel}, MGNO \cite{10.5555/3495724.3496291}, and Geo-FNO \cite{li2023fourier} have been proposed to enhance the perception of irregular geometric models by employing graph-based learning and geometric Fourier transforms. Several recent studies, such as 3D-GeoCA \cite{deng2024geometryguided} and Geom-DeepONet \cite{HE2024117130}, further enhance this capability by introducing techniques from the field of point cloud processing.

While operator learning methods demonstrate generalization ability in the face of variations in differential operators, challenges arise when dealing with other complex local variables such as geometries or boundary conditions. Without an efficient attention mechanism, these architectures often struggle to encode those intricate local features. It is worth mentioning that 3D-GeoCA \cite{deng2024geometryguided} uses a backbone structure to encode the geometric features and boundary conditions of the problem domain. In their experiments, this method was also integrated with Point-BERT \cite{pointbert} and other advanced Transformers. However, its main purpose is to enhance the capacity of the model to perceive complex geometric configurations, while the ability to generalize across multiple types of inputs remains unexplored.

\subsection{Transformer for PDEs}

Several past works have explored the possibility of applying the Transformer architecture to the field of AI for PDEs. In which, HT-NET \cite{Ma2022HTNetHC} takes advantage of the Swin Transformer \cite{liu2021Swin} and the multigrid method to enhance the ability of multi-scale spatial perception. OFormer \cite{Li2022TransformerFP} and IPOT \cite{10.1609/aaai.v38i1.27766} use Transformer to learn operators from meshes with varying resolutions. Recent studies, GNOT \cite{hao2023gnot} and UPT \cite{alkin2024universal}, attempted to enhance the multi-input generalization and multi-scale ability of the model. In addition, there are several studies devoted to improving the attention mechanism and increasing the accuracy of neural networks in fitting PDEs or operators \cite{cao2021choosetransformerfouriergalerkin,li2024scalable,xiao2023improved}.

The vast majority of the above models are trained using labeled data, which are extremely expensive in real industrial scenarios. While PINNs have demonstrated considerable potential in reducing data dependency, these complex networks have more complex gradients and are harder to converge when using PDEs as part of the loss function \cite{unknown}. In this regard, PIT \cite{santos2023physicsinformed} and PinnsFormer \cite{zhao2024pinnsformer} explored direct training through physics-informed methods and have been successful in some small-scale cases. Nevertheless, they did not consider generalization to more complex input conditions.

In addition, most of these studies apply the attention mechanism to the whole domain and directly embed each sample point as a token. When the complexity of the geometric model increases, the number of sample points or meshes required may reach tens or even hundreds of thousands, making the sequence length unacceptable for the global attention mechanism. Transolver \cite{wu2024Transolver} proposed a new physics-attention mechanism for segmenting geometric meshes to solve this problem, making it possible to solve complex geometric models under a fixed sequence length. Nevertheless, it relies on generating a weight map for each slice through a feedforward subnetwork, which lack global receptive fields to capture diverse input conditions. Therefore, a well-trained slicing subnetwork can only be effective for dynamic inputs within limited ranges. To address these varying conditions in a more universal manner, a predefined, non-learned slicing strategy should be considered.

\section{Problem Setup}

We consider the following PDE in a general form:
\begin{equation}
	\mathcal{N}(u;\lambda)=0, \quad x\in\Omega,t\in[0,T],
\end{equation}
where $u(x,t)$ is the expected value to be solved; $\mathcal{N}(u;\lambda)$ is the differential operator that describes the PDE, which is parameterized by physical parameters $\lambda$; $\Omega$ is the spatial domain that depends on the geometry; $T$ is the max time of the field.

For the complete definition of a problem instance, the PDE should also have the following initial conditions: 
\begin{equation}
		u(x,0) = g(x), \quad x \in \Omega,
\end{equation}
and different types of boundary conditions:
\begin{equation}
	\begin{aligned}
		u &= d(x,t), \quad &x &\in \partial\Omega_{\text{d}},t\in[0,T],\\
		\partial u/\partial x &= n(x,t), \quad &x &\in \partial\Omega_{\text{n}},t\in[0,T],\\
		a(\partial u/\partial x) + bu &= r(x,t), \quad &x &\in \partial\Omega_{\text{r}},t\in[0,T],
	\end{aligned}
\end{equation}
where $\partial\Omega$ represents the boundary of the problem domain, and $d$, $n$, $r$ are the Dirichlet, Neumann, and Robin boundary conditions, respectively.

To develop an ideal large-scale universal model for PDE solving, it is necessary to design a network characterized by the following mapping relationship:
\begin{equation}
	f(x,t,\lambda,\Omega,g,d,n,r; \theta)\rightarrow u,
\end{equation}
where $\theta$ are the parameters of the model, the domain $\Omega$ and initial/boundary conditions $g$, $d$, $n$, $r$ are highly complex and locally varying. The mapping $f$ is required to accurately solve a set of instances with the above variables, which represents a general class of problems in the engineering field.

\begin{figure*}[thpb]
	\centering
	\begin{subfigure}[b]{0.35\textwidth}
		\centering
		\includegraphics[scale=0.43, trim=27 15 25 15, clip]{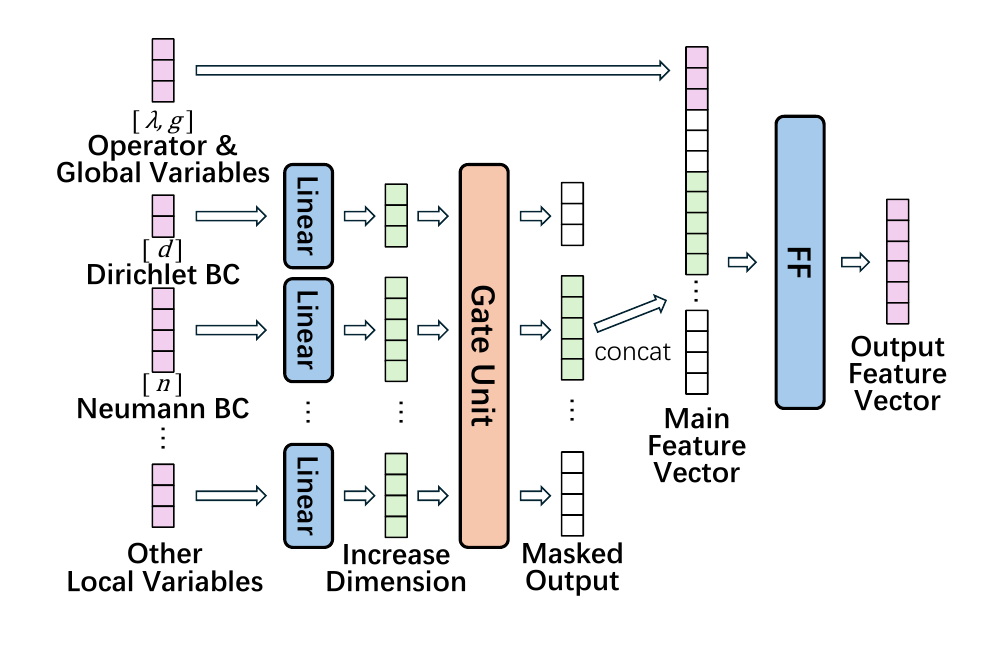}
		\caption{Pipeline of the GCE layer}
		\label{fig:sub1}
	\end{subfigure}
	\begin{subfigure}[b]{0.64\textwidth}
		\centering
		\includegraphics[scale=0.43, trim=17 15 27 20, clip]{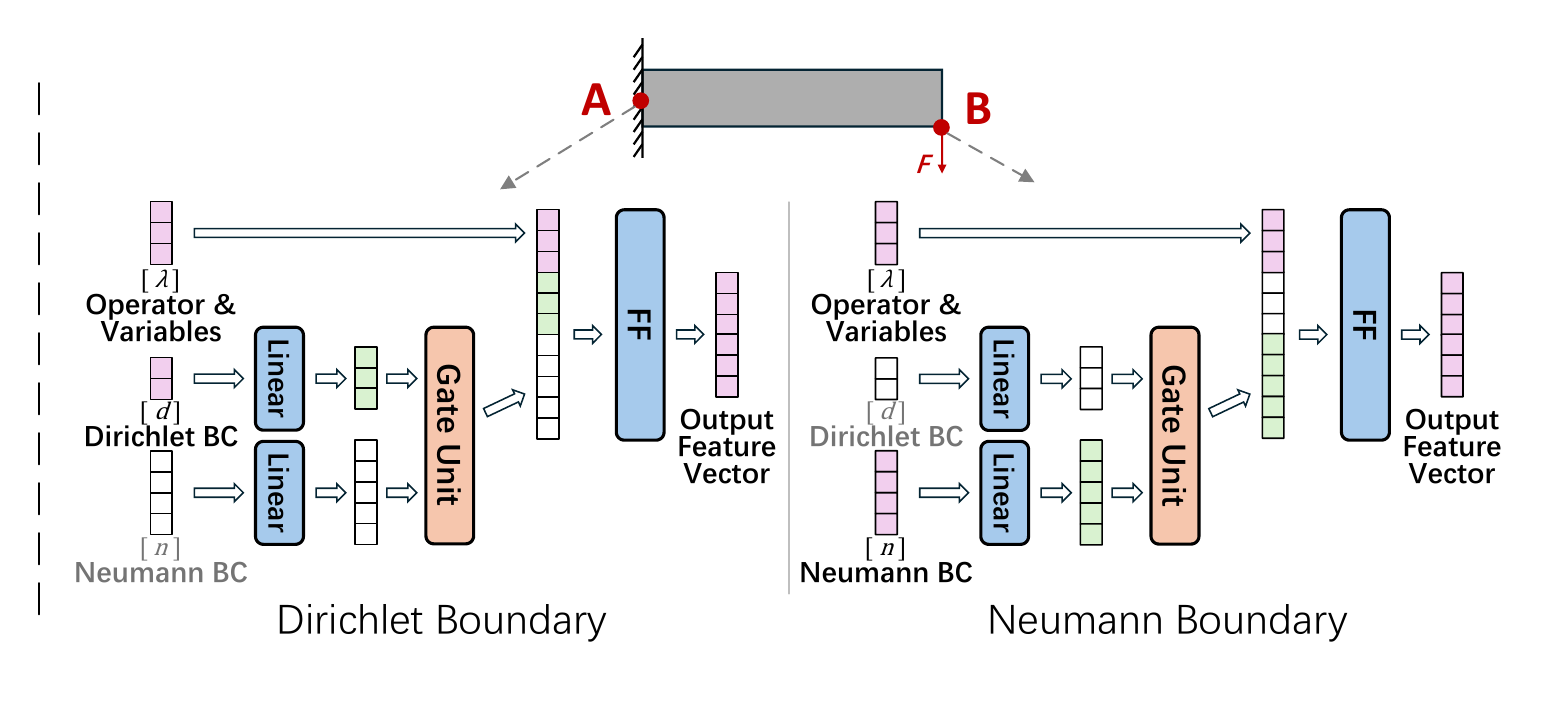}
		\caption{An example of the GCE layer to process different types of boundary conditions}
		\label{fig:sub2}
	\end{subfigure}
	
	\caption{Pipeline of the GCE layer and an example. The white square in the vector indicates that the value at that position is missing. Through dimension expansion and gating, this module can effectively extract features from different dimensions of input variables.}
	\label{fig:main}
\end{figure*}

\section{Our Method}

\subsection{Overall Architecture}

Given the complexity of geometric features, a common approach is to discretize them into point cloud or unstructured mesh for feature extraction. In MMET, we employ two discretization strategies for the instance under consideration, as shown in Figure \ref{arch1}. First, we adopt the same mesh generation technique as used in classical FEM \cite{Okereke2018}. The generated mesh serves as the input pseudo-sequence of the encoder to extract the global features. For the region of interest, the domain is further discretized into points to be queried, and is subsequently fed into the decoder.

In the encoder part of Figure \ref{arch1}, the input mesh pseudo-sequence comprises not only the spatial-temporal coordinates $(x,t)$, but also the sampling values of the differential operator $\mathcal{N}$, initial/boundary conditions $g$, $d$, $n$, $r$, and other necessary variables at this point. The coordinates of each point are directly fed into the positional encoding layer as position indices, and this layer can adopt a learnable feedforward network or fixed positional encoding as shown in PIT \cite{santos2023physicsinformed}. To handle complex input variables or functions, the innovative GCE layer is employed for effective embedding, which will be described in detail in Section \ref{gce}. Subsequently, the sequence is reserialized using a Hilbert curve for spatial coherence and divided into patches of a specified size. Each patch is transformed into a token via a linear layer, which will be described in Section \ref{pe}. The resulting token sequence is fed into the encoder for feature encoding.

In the decoder part of Figure \ref{arch1}, to enable querying at arbitrary resolutions, each query point is directly treated as a token and fed into the decoder. The self-attention layer typically used in the decoder of the classical Transformer \cite{vaswani2023attentionneed} is removed to ensure that the output is not influenced by the construction of the query sequence itself. Therefore, the query point sequence only performs cross-attention with the output feature sequence of the encoder. This design offers significant flexibility, allowing query points to be sampled in any quantity and at any location within the domain $\Omega$. Furthermore, if large number of points must be queried simultaneously or hardware resources are limited, the query sequence can be divided into batches and processed sequentially without affecting the final result.

In our model, all activation functions are replaced by the wavelet function proposed in PinnsFormer \cite{zhao2024pinnsformer} for better accuracy. The attention layer can be configured as classical dot-product attention \cite{vaswani2023attentionneed}, Galerkin attention \cite{cao2021choosetransformerfouriergalerkin}, or linear attention \cite{hao2023gnot}, depending on the specific requirements.

\subsection{Gated Condition Embedding}
\label{gce}

The previous approach GNOT \cite{hao2023gnot} uses an architecture similar to Mixture-of-Experts (MoE) approach \cite{45929,Fedus2021SwitchTS} to encode different features, which may lead to extremely long input sequences caused by repeated encoding of different features at the same point. The GCE layer is a novel module we designed specifically to address this deficiency. As shown in Figure \ref{fig:sub1}, we embed different input functions in units of spatio-temporal points $(x,t)$ rather than the sample values sequence of each function. Under this embedding approach, samples of different boundary conditions $d$, $n$, and $r$ exhibit different dimensional properties depending on whether the boundary point corresponds to Dirichlet, Neumann, or Robin types.

As an example, in Figure \ref{fig:sub2}, we consider two boundary points $A$ and $B$ with Dirichlet and Neumann boundary conditions $d_A(x_A,t_A)$ and $n_B(x_B,t_B)$, respectively. The input vector of point $A$ only needs to contain its sampled value $d_A$, while point $B$ may need to contain not only sampled value $n_B$, but also the normal vector $\Vec{n}$ of that boundary. However, neural networks require input vectors to have fixed dimensions. Naively masking missing features with zeros introduces ambiguity, as a fully connected network cannot distinguish between an input value that is genuinely zero and one representing missing data.

The GCE layer resolves this issue by mapping each type of input variable to a higher-dimensional space using a linear layer. After training, the original feature dimensions form a hyperplane in this higher-dimensional space that does not intersect the origin. If the input vectors of points $A$ and $B$ are expressed as vectors $\Vec{d}_A \in \mathbb{R}^a$, $\Vec{n}_B \in \mathbb{R}^b$, the outputs of the proposed linear layers are given as:
\begin{equation}
	\begin{aligned}
		\Vec{h}_A&=W_d\cdot\Vec{d}_A + \Vec{b}_d, \quad & \Vec{h}_A\in \mathbb{R}^{a+1},\\
        \Vec{h}_B&=W_n\cdot\Vec{n}_B + \Vec{b}_n, \quad & \Vec{h}_B\in \mathbb{R}^{b+1},
	\end{aligned}
\end{equation}
where $W_d \in \mathbb{R}^{(a+1) \times a}$ and $W_n \in \mathbb{R}^{(b+1) \times b}$ are the weight matrices of the linear layers for Dirichlet and Neumann boundary condition inputs, respectively, and they are independent of each other; $\Vec{b}_d \in \mathbb{R}^{a+1}$ and $\Vec{b}_n \in \mathbb{R}^{b+1}$ are the bias vectors of these layers.

A gate unit is then used to control whether the mapped features $\Vec{h}_A$ and $\Vec{h}_B$ should be integrated into the embedding vector. If this feature is missing, the value at that position in the embedding vector will be set to zero, representing the origin in the high-dimensional space. Importantly, this gate unit is parameter-free and relies solely on the presence or absence of the corresponding variables for a given point. Therefore, the embedding vectors of points $A$ and $B$ can be expressed as:
\begin{equation}
	\begin{aligned}
		\Vec{e}_A&=\text{MLP}[\text{concat}(\lambda, g, \Vec{h}_A, \Vec{o}_A)], \quad & \Vec{e}_A\in \mathbb{R}^{c},\\
        \Vec{e}_B&=\text{MLP}[\text{concat}(\lambda, g, \Vec{o}_B, \Vec{h}_B)], \quad & \Vec{e}_B\in \mathbb{R}^{c},
	\end{aligned}
\end{equation}
where $\Vec{o}_A \in \mathbb{R}^{b+1}$ and $\Vec{o}_B \in \mathbb{R}^{a+1}$ are zero vectors masked by the gate unit, which are used to pad missing inputs.

This design allows the model to reliably distinguish between missing inputs and inputs with a value of zero, ensuring a more robust and interpretable representation.

\subsection{Reserialization and Patch Embedding}
\label{pe}

Patching the mesh sequence to a larger slice not only reduces its overall length but also enhances the model's performance on complex instances, as directly applying attention mechanisms to a large number of mesh points often hinders the model's ability to capture meaningful relationships \cite{wu2022flowformer}. To ensure generalization to different input conditions, we avoid using any learnable patch method such as those proposed in Transolver \cite{wu2024Transolver}.

Inspired by the work in point cloud processing, PTv3 \cite{wu2024ptv3}, we first reserialize the mesh points using the Hilbert curve \cite{hilbert1935stetige}, which is a type of space-filling curve designed to map high-dimensional discrete spaces into one-dimensional sequences while preserving spatial coherence. Compared with alternatives like Z-order curves or random raster scans, Hilbert ordering more consistently maps nearby nodes in 2D/3D space to adjacent positions in the serialized sequence, making it well-suited for downstream patching. This improves the likelihood that spatially related features are grouped within the same patch, which is especially beneficial for attention-based models operating under fixed memory budgets.

\begin{figure}[thpb]
	\centering
	\begin{subfigure}[b]{0.5\textwidth}
		\centering
		\includegraphics[width=0.9\linewidth, trim=10 10 10 10, clip]{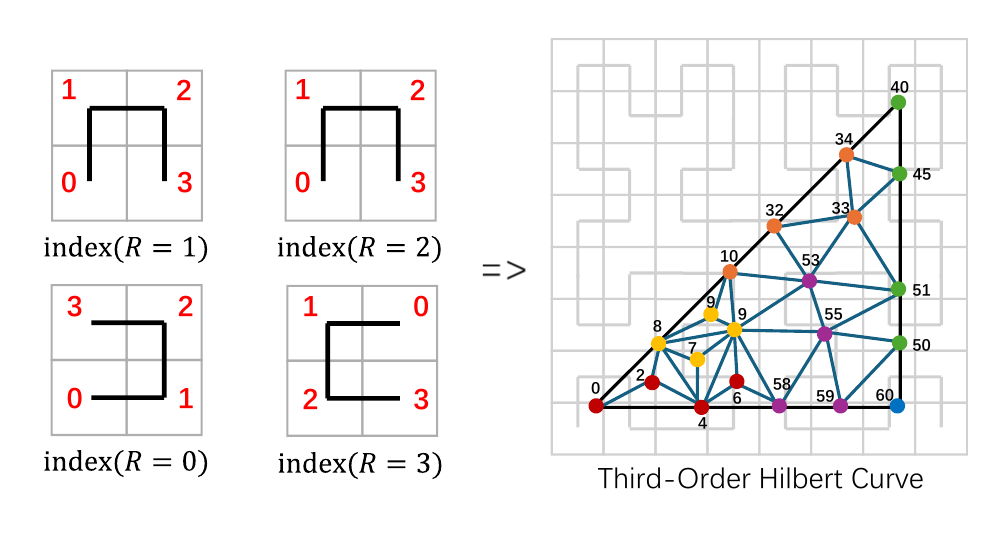}
		\caption{Reserialize the mesh using Hilbert curve}
		\label{fig:sub23}
	\end{subfigure}
	\begin{subfigure}[b]{0.5\textwidth}
		\centering
		\includegraphics[width=0.9\linewidth, trim=10 10 10 10, clip]{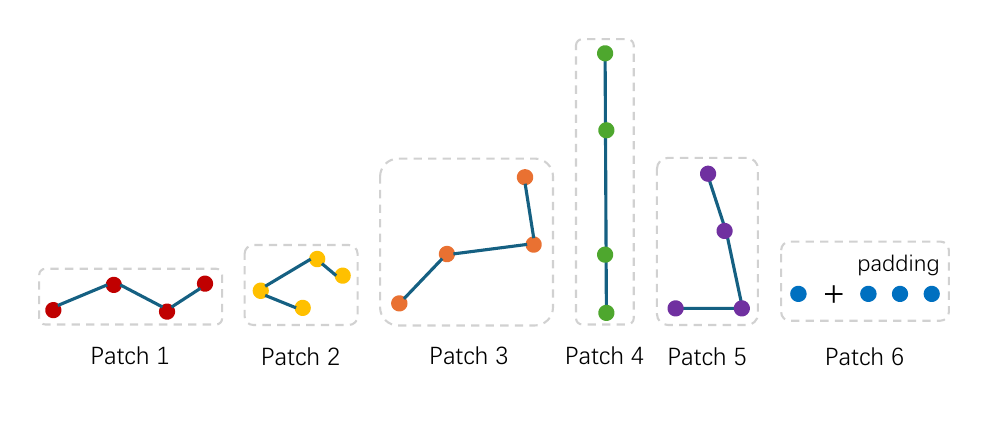}
		\caption{Patching the mesh nodes as 1D sequence}
		\label{fig:sub24}
	\end{subfigure}
	
	\caption{Reserialize and patching of the mesh. Points with same color indicate the same patch. Regions with denser nodes are assigned more patches, as patch 1 and patch 2.}
	\label{fig:main1}
\end{figure}

As shown in Figure \ref{fig:main1}, we take a 2D mesh as an example. Consider a node $(x, y)$ within this unstructured mesh, linearly mapped into a square where each dimension is discretized into $2^n$ units. The Hilbert curve recursively subdivides the square into four subregions, with the number of recursion determined by the order number $n$. Each subregion at a given order $i$ is sequentially assigned a code $b_i\in[0,3]$, which can be expressed as:
\begin{equation}
	b_i=\text{index}(x,y,R_i),\quad i\in [0,n],
\end{equation}
where $R_i$ represents the rotation direction at order $i$, which depends on the starting orientation of the recursion. The $\text{index}()$ function is visualized in Figure \ref{fig:sub23}, dynamically adjusting the rotation direction $R_i$ to ensure that spatially adjacent coordinates maintain locality in the serialized sequence. The Hilbert code $e$ for each coordinate can be computed as:
\begin{equation}
	e=\sum_{i=0}^{n-1}b_i\times2^{2i}.
\end{equation}

Each node in the original mesh is sorted according to their Hilbert code $e$ to form a new 1D sequence. This sequence is then patched according to a fixed size. The vectors within each patch are concatenated and passed through a linear layer to extract patch features, which serve as individual tokens for the encoder.

This approach further benefits from FEM's adaptive meshing: regions with sharp gradients are a priori assigned denser meshes. By fixing the patch size, these regions generate more tokens, allowing the attention mechanism to prioritize critical areas. This design enhances MMET's capability in multi-scale scenarios while preserving accuracy for complex geometries.

\section{Experiment and Result}

\subsection{Benchmark Comparisons}

\begin{table*}
    \centering
    \begin{small}
	    \begin{tabular}{l|cc|ccc|cc}
	        \toprule
	        \multirow{2}{*}{\textbf{Dataset}}  & \multicolumn{2}{c|}{\textbf{Property}} & \multicolumn{5}{c}{\textbf{Relative $L_2$ Average Error}}\\
	        & \textbf{Challenge} & \textbf{Variable} & \textbf{PinnsFormer} & \textbf{Transolver} & \textbf{GNOT} & \textbf{MMET} \small{(encoder-only)} & \textbf{MMET} \\
	        \midrule
	        Poisson             & $-$           & $u$           & \textbf{4.90e-3} & 7.19e-2 & 2.73e-2 & 1.12e-2 & 5.65e-2 \\
	        Darcy Flow	        & $-$           & $p$           & 3.69e-2 & 7.87e-2 & 3.23e-1 & 3.83e-2 & \textbf{3.34e-2} \\
	        Shape-Net Car       & $A$           & $v$           & 4.88e-2 & 3.29e-2$^*$ & 2.07e-2$^*$ & \textbf{1.93e-2} & 4.35e-2 \\
                                &               & $p$           & 1.21e-1 & 7.98e-2$^*$ & 7.45e-2$^*$ & \textbf{7.38e-2} & 1.11e-1 \\
	        Heat2d              & $A$, $B$      & $T$           & - & - & 4.13e-2$^*$ & 3.42e-2 & \textbf{3.36e-2} \\
	        Beam2d              & $B$, $C$      & $u$           & - & - & 3.37e-1 & - & \textbf{4.64e-2} \\
	                            &               & $v$           & - & - & 8.71e-2 & - & \textbf{7.46e-3} \\
	                            &               & $\sigma_\text{v}$    & - & - & 1.08e-2 & - & \textbf{1.16e-3} \\
	        HeatSink2d          & $A$, $B$, $C$ & $T$           & - & - & 1.33e-1 & - & \textbf{1.18e-1}    \\
	        \bottomrule
	    \end{tabular}
	\end{small}
    \caption{Performance comparison on our selected datasets. Where "-" indicates that the model does not support solving such problems, and results of other models with "*" indicate that these results are from their original paper. A lower error value indicates that the model performs better. In the Beam2d dataset, $\sigma_\text{v}$ represents the von Mises yield stress, which is calculated by normal stresses $\sigma_x$, $\sigma_y$, and shear stress $\tau_{xy}$.}
    \label{tab:booktabs}
\end{table*}

In this section, we conduct benchmark experiments and compare our model to demonstrate its advantages over previous methods.

\textbf{Experimental Setting.} We select six representative datasets spanning elasticity, fluid mechanics, and thermodynamics. Here, $A$ indicates a dataset with multiple geometric shapes, $B$ indicates datasets with various boundary conditions, and $C$ indicates datasets with multi-scale query resolution. The following is a brief introduction.

\begin{itemize}
\item Poisson: A simple 2D Poisson problem that is mainly used to verify the effectiveness of each baseline under physics-informed training.

\item Darcy Flow \cite{pdebench}: The 2D Darcy flow problem from the PDEBench datasets, aims to learn the mapping from spatially varying diffusion coefficients to pressure distributions.

\item Shape-Net Car ($A$) \cite{10.1145/3197517.3201325}: A multi-scale 3D aerodynamics problem used to predict the flow velocity $v$ and surface pressure $p$ of different cars. This dataset includes complex geometric features.

\item Heat2d ($A$, $B$) \cite{hao2023gnot}: A multi-scale heat conduction problem used to predict the temperature field $T$ from input functions.

\item Beam2d ($B$, $C$) \cite{bai2023physics}: This dataset solves the stress $\sigma_x$, $\sigma_y$, $\tau_{xy}$, and displacement $u$, $v$ of a rectangular cantilever beam when the end is subjected to a bending moment. We modify this case so that the bending moment varies dynamically within a certain range.

\item HeatSink2d ($A$, $B$, $C$): A thermodynamic case designed to solve the internal temperature $T$ of a 2D heat sink with variable height and surface temperature. This case relies solely on PDEs for physics-informed training, and without using labeled data. \end{itemize}

Meanwhile, we selected another three recent Transformer-based SOTA methods as baselines, including PinnsFormer \cite{zhao2024pinnsformer}, Transolver \cite{wu2024Transolver}, and GNOT \cite{hao2023gnot}. For MMET, we also tested the encoder-only configuration on some of the datasets that do not need for multi-scale queries.

For all models, we employ AdamW \cite{loshchilov2018decoupled} or L-BFGS \cite{10.5555/3112655.3112866} optimizers for combinatorial optimization. The model is trained on two NVIDIA H20 GPUs. The relative $L_2$ error is used to evaluate all models, which has also been widely applied in previous works \cite{wu2024Transolver,xiao2023improved,hao2023gnot}. The error is computed as follows:
\begin{equation}
e = \frac{1}{N}\sum^N_{i=1}\frac{\| u_i - u_i' \|_2}{\| u \|_2},
\end{equation}
where $u_i$ is the prediction value output by the model, $u_i'$ is the ground truth, and $N$ is the number of data in the test set.

For the Poisson, Shape-Net Car and Darcy Flow dataset, since there are no multiple boundary condition inputs to embed, we therefore replace the GCE layer by a simple feedforward network. For the HeatSink2d dataset, the error is expected to be larger than for others because training relies solely on PDEs, while the ground truth from Ansys Workbench (via FEM) also contains approximation errors.

\begin{figure}[thpb]
	\centering
	\includegraphics[width=0.32\linewidth, trim=12 7 12 10, clip]{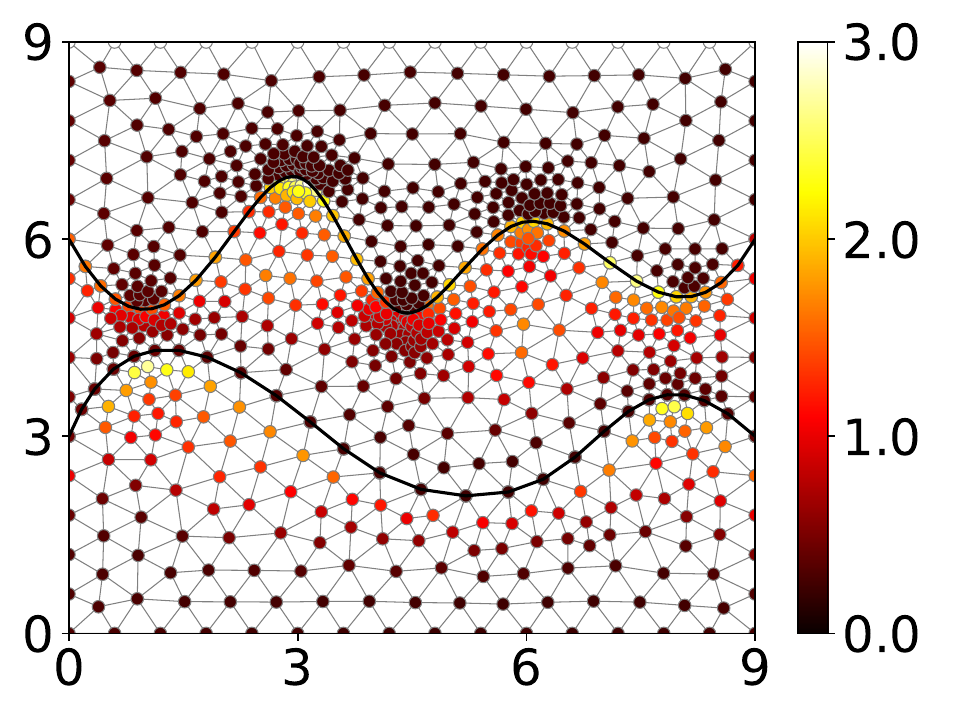}
	\includegraphics[width=0.32\linewidth, trim=12 7 12 10, clip]{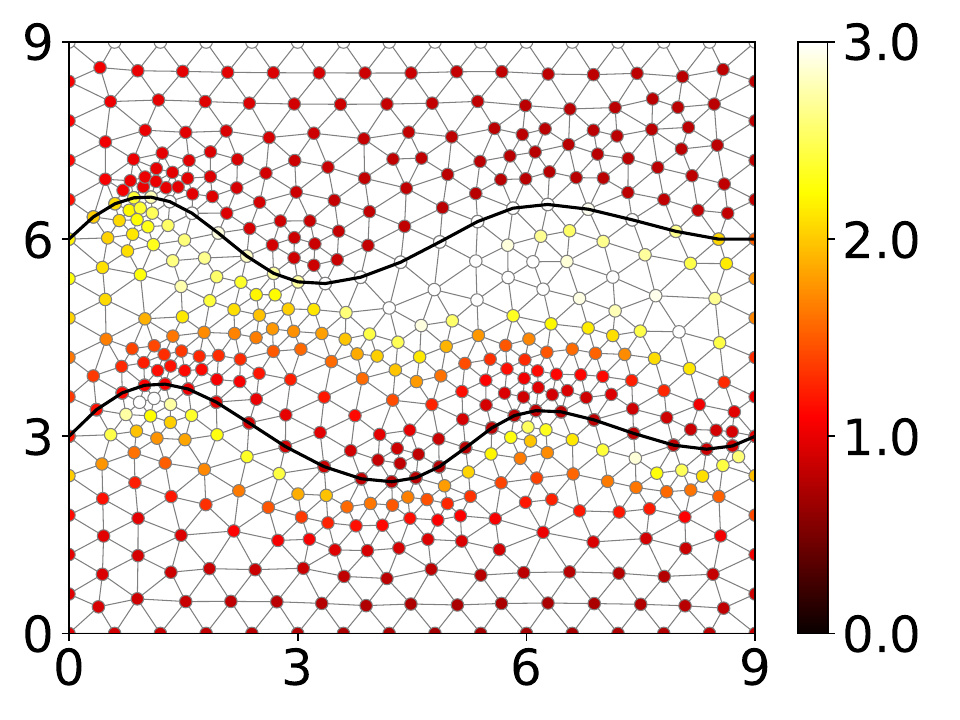}
	\includegraphics[width=0.32\linewidth, trim=12 7 12 10, clip]{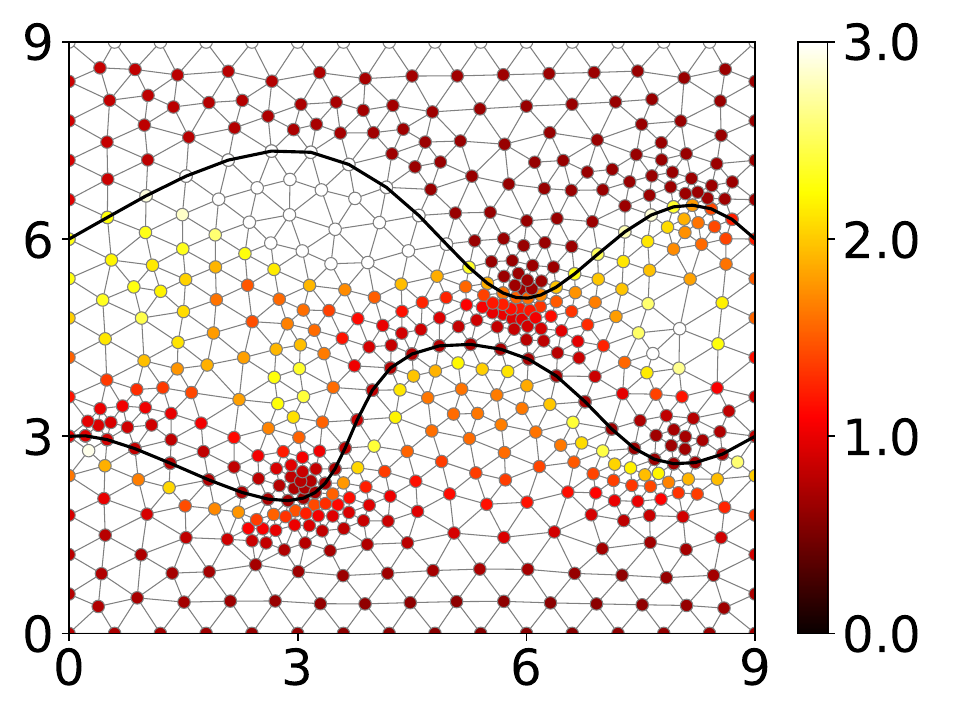}
	\includegraphics[width=0.32\linewidth, trim=12 7 12 10, clip]{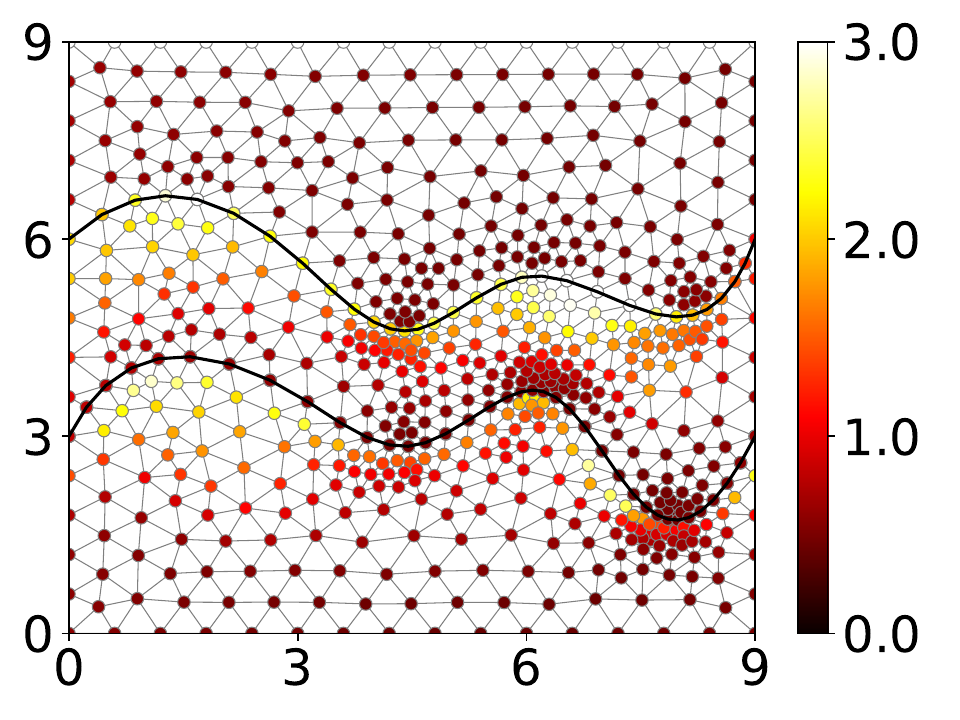}
	\includegraphics[width=0.32\linewidth, trim=12 7 12 10, clip]{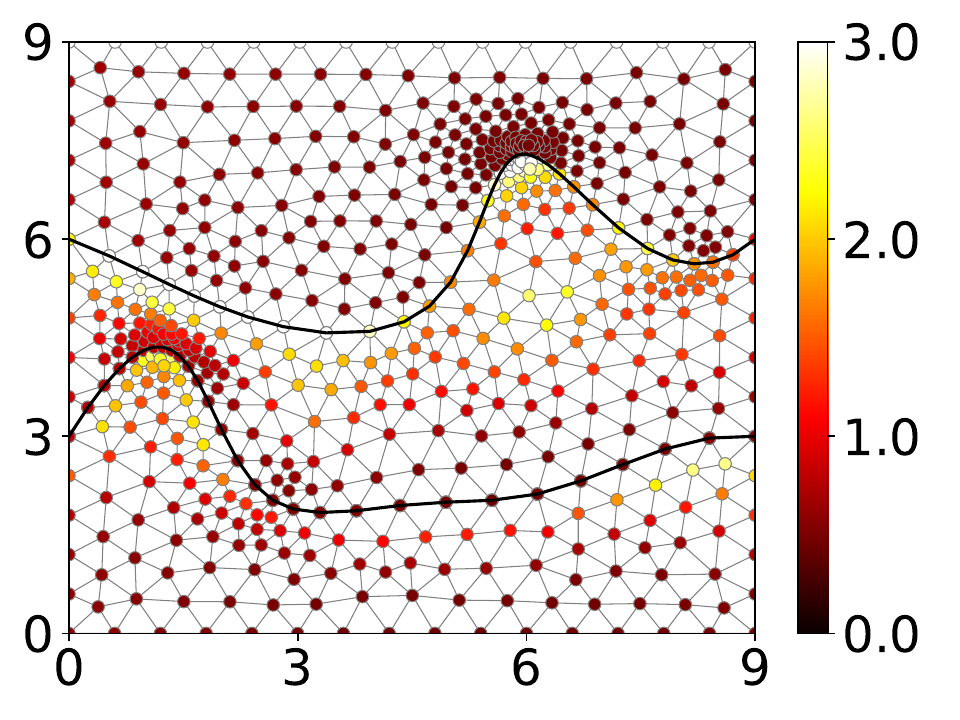}
	\includegraphics[width=0.32\linewidth, trim=12 7 12 10, clip]{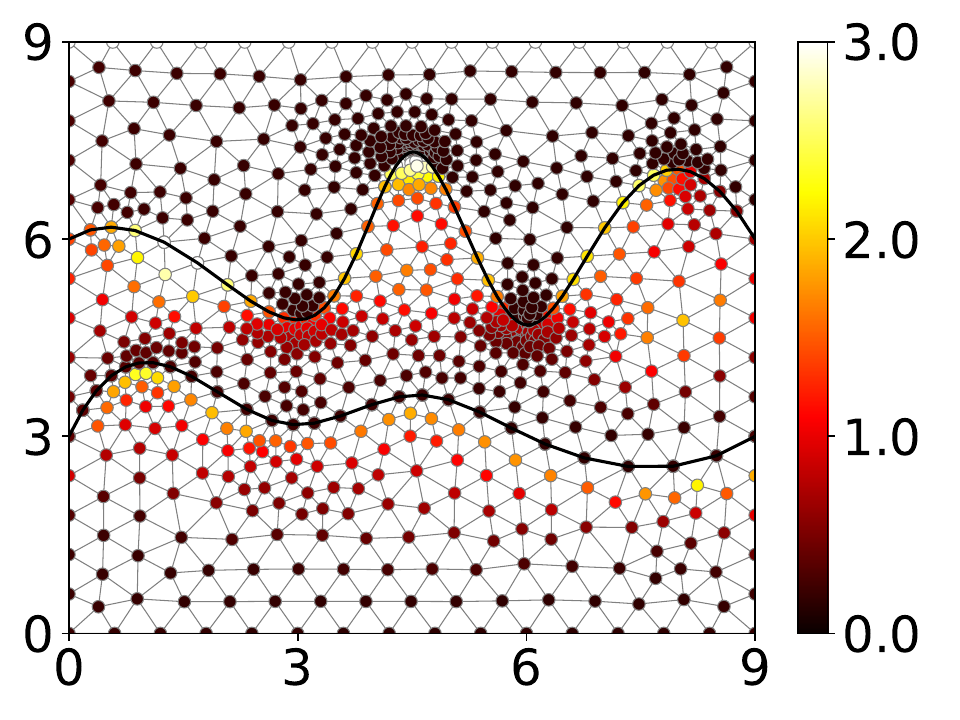}
	\caption{The attention heatmap of the decoder on the Heat2d dataset, under varying boundary conditions and geometries, adaptively focuses on critical regions such as geometric boundaries or gradients, confirming the spatial interpretability of MMET.}
	\label{result34}
\end{figure}

\textbf{Result and Discussion.} The results in Table \ref{tab:booktabs} demonstrate that our model consistently outperforms selected SOTA baselines in nearly all complex datasets, highlighting the effectiveness of MMET in addressing large-scale and dynamic physical fields. In contrast, PinnsFormer and Transolver lack effective encoding mechanisms for boundary conditions, rendering them incapable of handling problems with multiple input conditions. It is worth noting that on the Heat2d dataset designed by GNOT's team (which is characterized by multi-input, multi-geometry, and multi-scale), our model shows higher accuracy than GNOT. This further validates the capability of MMET in tackling intricate and diverse scenarios.

However, we also observe that the encoder-decoder structure in complete MMET converges more slowly and is sensitive to large learning rates. Therefore, for tasks that do not require arbitrary resolution queries, such as Shape-Net Car, an encoder-only variant is recommended to improve training speed, stability, and accuracy.

\subsection{Multi-Scale Query Experiment}

\begin{figure}[thpb]
    \centering
        \begin{tabular}{@{}m{0.05\linewidth}@{}m{0.95\linewidth}@{}}
            & \begin{tabular}{@{}m{0.45\linewidth}@{}m{0.55\linewidth}@{}}\centering \footnotesize{x-Displacement $u$} & \centering \footnotesize{Absolute Error}\end{tabular} \\
            \footnotesize{(a)} & \includegraphics[scale=0.26]{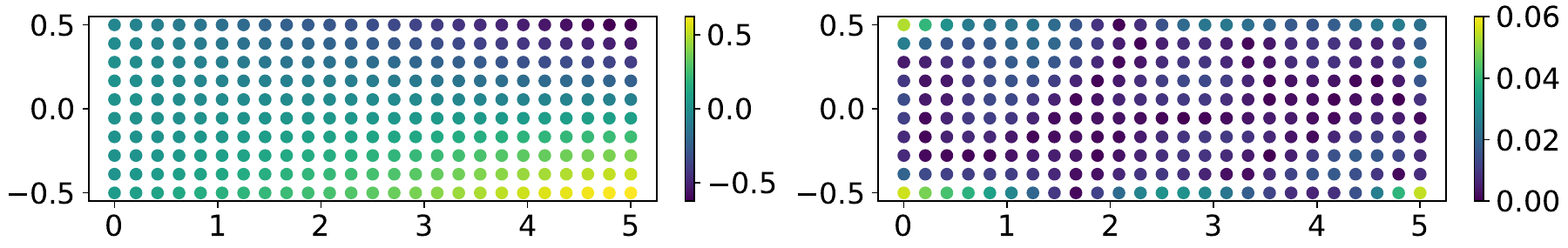}\\
            \footnotesize{(b)} & \includegraphics[scale=0.26]{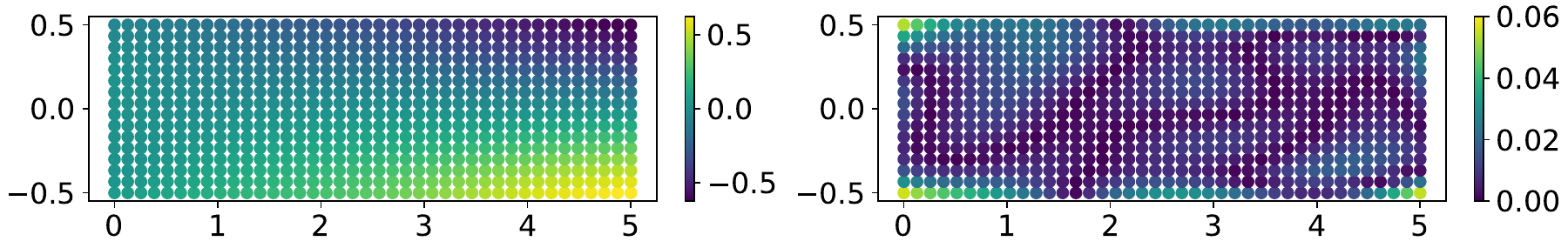}\\
            \footnotesize{(c)} & \includegraphics[scale=0.26]{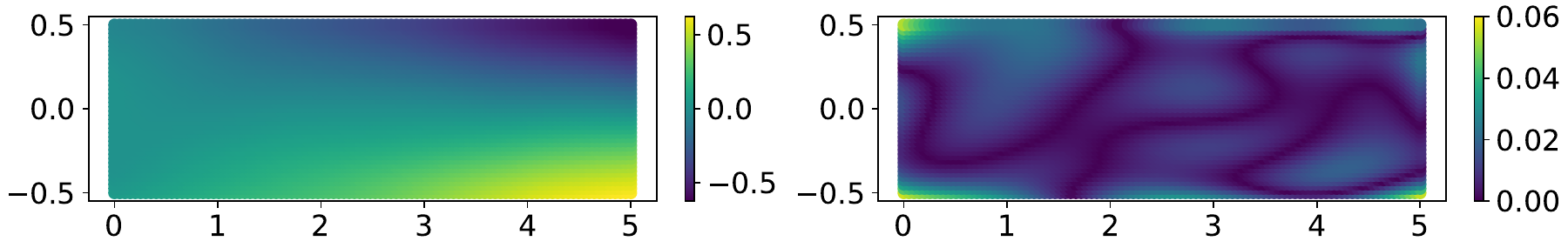}\\
            \footnotesize{(d)} & \includegraphics[scale=0.26]{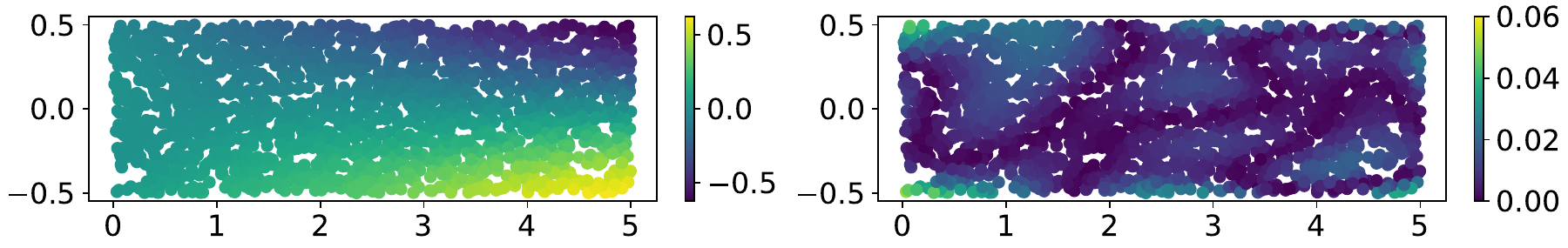}\\
        \end{tabular}
    \caption{Results of x-axis displacement $u$ queries under different resolutions. Where (a) is the result under $[25\times10]$ resolution; (b) is the result under $[40\times16]$ resolution; (c) is the result under $[100\times40]$ resolution; (d) is the result under $2000$ completely random points.}
    \label{result}
\end{figure}

We designed a supplementary experiment to verify the zero-shot query ability of our model at different resolutions. For the Beam2d dataset, in the training stage, our model uses an unstructured mesh with $5404$ nodes generated by Ansys Workbench. The query sequence is a regular point matrix with a resolution of $[50\times20]$. In the inference stage, we keep the mesh input the same as in the training stage and test the performance of the query points at four different resolutions, as shown in Figure \ref{result}.

Experimental results indicate that decoupling the mesh from the query sequence enables our model to maintain fully consistent error performance across varying query resolutions. This demonstrates the distinct fitting capability of MMET within the continuous function space, comparable to that of traditional fully connected networks. In contrast, other Transformer-based models suffer from accuracy fluctuations as the input sequence length changes.

\subsection{Ablation Experiment}

In this section, we evaluate the impact of disabling the GCE layer and modifying the patch size in different datasets, and analyzing their effects on the performance of our model.

\textbf{GCE Layer.} To assess the significance of the GCE layer, we designed an experiment introducing ambiguity in boundary representation. This experiment is modified from HeatSink2d dataset, where a boundary is represented as a heat flux (Neumann) boundary in part of the dataset and as a zero-value temperature (Dirichlet) boundary in the remainder. If a zero pad is used to pad the missing inputs, then the sampled characteristics of the two parts on this boundary would be exactly the same. We evaluated the performance of the model using three embedding approaches: a simple MLP, an MLP with an additional input node to encode the boundary type, and our GCE layer. The results of this experiment are presented in Table \ref{tab:gce}.

\begin{table}
    \centering
	\begin{small}
	    \begin{tabular}{l|ccc}
	        \toprule
	        \textbf{Dataset Part} & \textbf{MLP} & \makecell{\textbf{MLP +} \\ \textbf{Boundary Type}} & \textbf{GCE}\\
	        \midrule
	        HeatSink2d -D & 3.90e-1 & 2.42e-2 & \textbf{1.52e-2} \\
	        HeatSink2d -N & 2.61e-1 & 2.11e-2 & \textbf{9.86e-3} \\
	        \bottomrule
	    \end{tabular}
	\end{small}
    \caption{Relative $L_2$ error under different types of embedding layers. Where "-D" represents the part of the data in the dataset with temperature (Dirichlet) boundaries, and "-N" represents the part of the data with heat flux (Neumann) boundaries.}
    \label{tab:gce}
\end{table}

It can be seen that using a simple MLP layer results in complete failure, with errors exceeding $20\%$ in both parts of the dataset. This is because the exact same input makes the MLP unable to effectively distinguish between different types of boundary conditions. Introducing a boundary type indicator into the input indeed enables the model to learn effectively. Nonetheless, our GCE layer achieves significantly higher accuracy. This result shows the superior performance of our method in resolving input ambiguities and dynamically encodes boundary conditions with different dimensions.

\begin{figure}[t]
	\centering
        \begin{tabular}{@{}m{0.5\linewidth}@{}m{0.5\linewidth}@{}}
            \multicolumn{2}{c}{\includegraphics[width=0.55\linewidth, trim=15 20 25 20, clip]{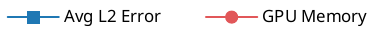}} \\
            \includegraphics[width=1\linewidth, trim=10 10 0 10, clip]{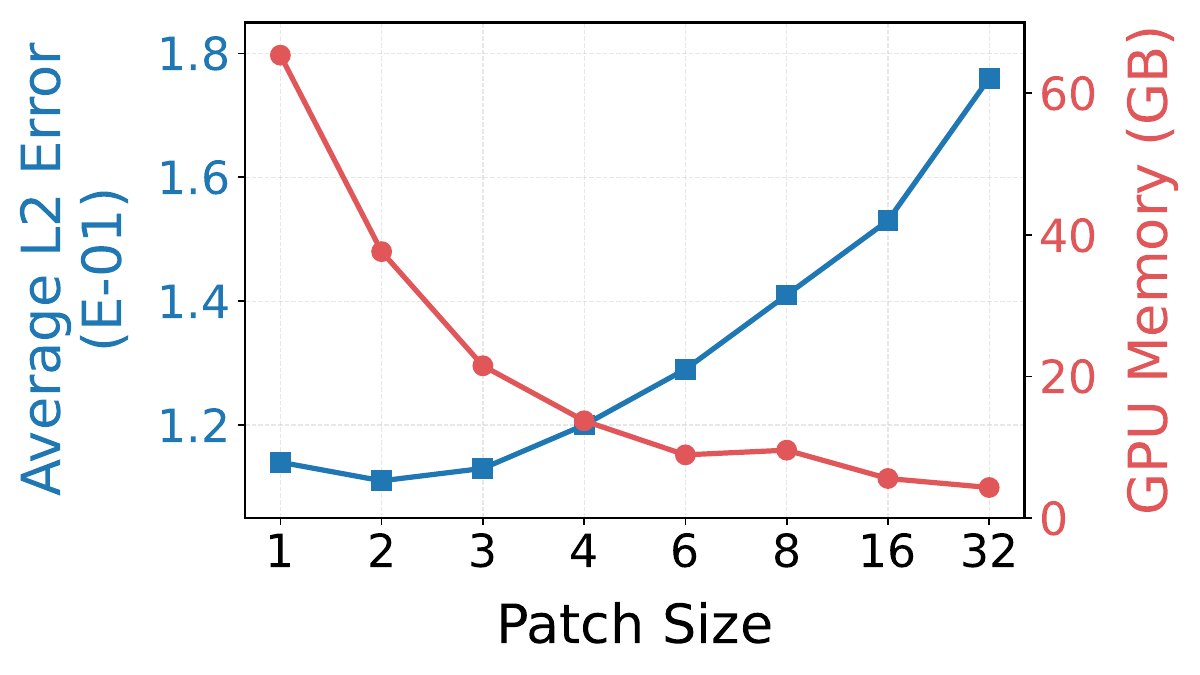} & \includegraphics[width=1\linewidth, trim=0 10 10 10, clip]{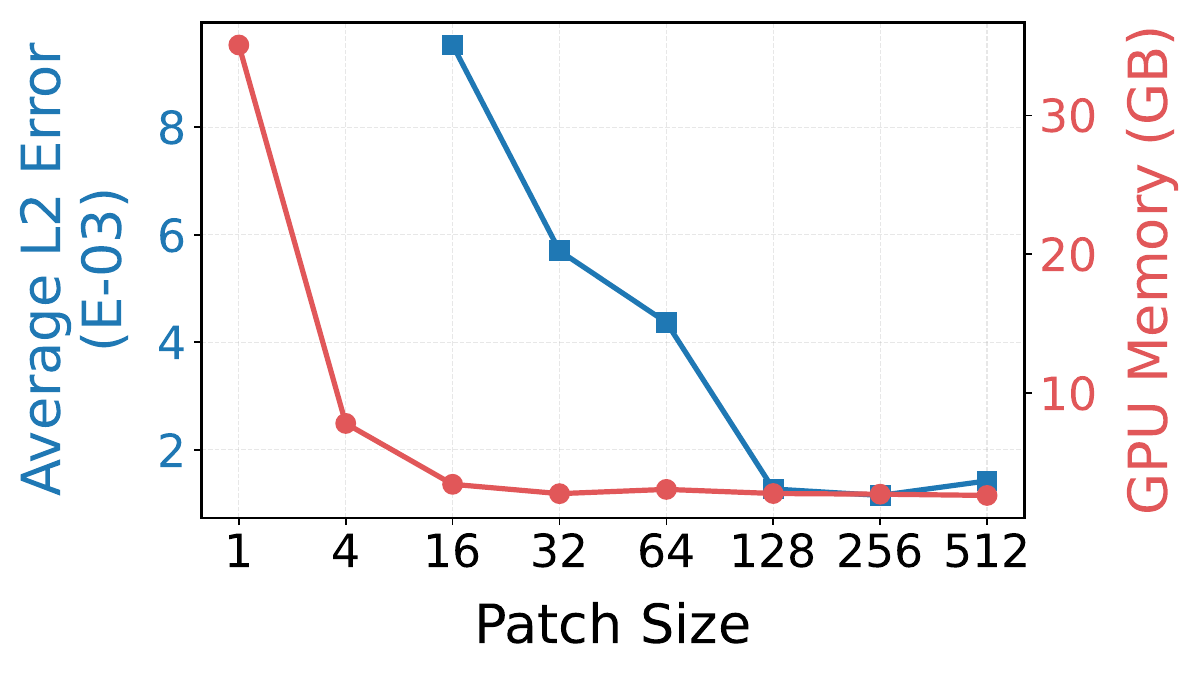}\\
            \makebox[\linewidth]{\footnotesize (a) Shape-Net Car} & \makebox[\linewidth]{\footnotesize (b) Beam2d} \\
        \end{tabular}
	\caption{The lowest relative $L_2$ error and GPU memory consumption for each dataset. Specifically, the GPU memory consumption is tested in the inference phase with a batch size of 50.}
	\label{resultxr}
\end{figure}

\textbf{Patch Embedding.} In the Shape-Net Car and Beam2d datasets, we dynamically adjust the patch size in a certain range to evaluate the effectiveness of our patching method. Notably, a patch size of 1 is equivalent to disabling patch embedding. To ensure a fair comparison, the model scale and training configuration are maintained consistently in all settings. The best accuracy and GPU memory consumption are shown in Figure \ref{resultxr}.

For the Shape-Net Car dataset, the experimental results indicate that our method surpasses the no-patch configuration when the patch size ranges from 2 and 3, while significantly reducing GPU memory consumption during training and inference. However, excessively large patch sizes lead to a noticeable decline in performance without obvious gains in efficiency. In contrast, for the Beam2d dataset, the model fails to converge with small patch sizes, indicating an inability to effectively extract information from a complex mesh. As the patch size increases, the convergence rate and final accuracy of the model improve progressively.

These findings highlight the effectiveness of our approach in handling complex meshes. The Shape-Net Car dataset, derived from 3D point clouds sampled uniformly across model surfaces, differs fundamentally from the Beam2d dataset, which consists of meshes generated by Ansys Workbench. The latter a priori incorporates physical information. This enables the model to prioritize regions with complex gradients more effectively. By employing fixed-size patches, our method implicitly assigns different weights to various regions of the geometry, thereby enhancing the efficiency and accuracy of the attention mechanism.

\section{Conclusion}

In this work, we proposed the MMET, a novel framework for real-time solving of generalized PDEs with multi-input and multi-scale resolution. The architecture incorporates key innovations, including the GCE layer and Hilbert curve-based reserialization and patching, enabling efficient handling of complex geometries and input conditions with scalability and accuracy. Experimental results across various benchmarks demonstrate the superior performance of MMET compared to SOTA methods, showcasing its potential for broad applications in complex physics-based simulations.

In the future, we plan to scale up the model, striving to establish it as a foundational pre-trained model for specific physics fields.

\appendix

\section*{Acknowledgments}
Yichen Luo is sponsored by the China Scholarship Council for PhD study at KTH Royal Institute of Technology, Sweden (No.202408320060). This work was partially supported by the National Natural Science Foundation of China (Grant No.72401232), the XJTLU External Research Fund under 2024-015, and LLL24ZZ-02-01 and LLL24ZZ-02-02 funded by the Liaoning Liaohe Laboratory.

\bibliographystyle{named}
\bibliography{ijcai25}

\newpage

\section*{Appendix A. Time Complexity}

We designed a supplementary experiment to quantify the time complexity of each algorithm, as shown in Table \ref{tab}. It can be seen that both the models with patch grouping (MMET and Transolver) have a significant advantage in time complexity. And due to the use of linear attention, the computation time of MMET is further slightly shorter than Transolver with same scale.

\begin{table}[h]
	\centering
	\begin{scriptsize}
		\begin{tabular}{l|ccc|c}
			\toprule
			\textbf{Phase} & \textbf{Transolver} & \textbf{PinnsFormer} & \textbf{GNOT} & \textbf{MMET}\\
			\midrule
			Training (s) & 934.36 & 3295.28 & 3013.88 & \textbf{896.27} \\
			Inference (ms) & 6.41 & 15.06 & 14.95 & \textbf{5.49} \\
			\bottomrule
		\end{tabular}
		\caption{Training and inference time consumption for all models on Shape-Net Car dataset. We use RTX-4090 GPU, and keep the same hyperparameters and model scale for all experiments.}
		\label{tab}
	\end{scriptsize}
\end{table}

\section*{Appendix B. Dataset Description}

\subsection*{Poisson}

Poisson equation is a kind of fundamental PDE commonly encountered in physics and engineering. In this case, we consider a two-dimensional Poisson equation on a unit square domain $\Omega = [0, 1] \times [0, 1]$:
\begin{equation}
	-\nabla^2 u(x, y) = f(x, y), \quad (x, y) \in \Omega,
\end{equation}
where $u(x, y)$ is the unknown solution and $f(x, y)$ is the source term. For this case, the source term is chosen as:
\begin{equation}
	f(x, y) = -2\pi^2 \sin(\pi x) \sin(\pi y).
\end{equation}

The boundary condition is specified as:
\begin{equation}
	u(x, y) = 0, \quad (x, y) \in \partial \Omega,
\end{equation}
where $\partial \Omega$ denotes the boundary of the domain. The analytical solution for this problem is:
\begin{equation}
        \label{eq1}
	u_{\text{true}}(x, y) = \sin(\pi x) \sin(\pi y).
\end{equation}

In this case, we use both PDE and labeled data to constrain convergence of the model. The PDE residual is defined as:
\begin{equation}
	r_{\text{PDE}}(x, y) = -\nabla^2 u(x, y) - f(x, y),
\end{equation}
where the Laplacian $\nabla^2 u(x, y)$ is computed using automatic differentiation tool:
\begin{equation}
	\nabla^2 u(x, y) = \frac{\partial^2 u}{\partial x^2} + \frac{\partial^2 u}{\partial y^2}.
\end{equation}

Therefore, the PDE residual loss is given by:
\begin{equation}
	\mathcal{L}_{\text{PDE}} = \sum r_{\text{PDE}}^2.
\end{equation}

The data loss is obtained from the analytical solution in Equation \ref{eq1}. It is defined as:
\begin{equation}
	\mathcal{L}_{\text{data}} = \sum (u-u_{true})^2.
\end{equation}

Therefore, the total loss is the sum of the PDE residual loss and the data loss:
\begin{equation}
	\mathcal{L} = \mathcal{L}_{\text{PDE}} + \mathcal{L}_{\text{data}}.
\end{equation}

For each of these models, the training input is a $[50\times 50]$ regular point matrix, and the test input is a $[100\times 100]$ matrix. The main purpose of this example is to evaluate the stability of the individual models under physics-informed training approach \cite{RAISSI2019686}, so we do not distinguish between training and test sets.

\subsection*{Darcy Flow}

The Darcy Flow dataset is a 2D dataset for steady-state subsurface flow problems within the PDEBench benchmark \cite{pdebench}. Its core features include:
\begin{itemize}
    \item Heterogeneous parameter field: The permeability field $\theta(s)$ exhibits spatial heterogeneity (see Equation (2) in the original paper), simulating the complex physical processes of real geological structures.
    \item Diverse boundary conditions: It provides non-periodic boundaries (e.g., Neumann/Dirichlet conditions), differing from the periodic boundary assumptions commonly found in scientific machine learning.
\end{itemize}

The dataset contains 10,000 samples with a spatial resolution of $128 \times 128$ (no temporal dimension). The data is stored in HDF5 format and can be downloaded and reproduced via the official PDEBench repository: \url{https://github.com/pdebench/PDEBench}.

\subsection*{Shape-Net Car}

Shape-Net Car \cite{10.1145/3197517.3201325} is a multi-scale 3D aerodynamics problem, which includes complex geometric features. This dataset contains two parts to predict the drag coefficient value $c_d$, surrounding velocity $v$ and surface pressure $p$ of 3D cars of different shapes, respectively. By collecting the wind pressure and wind speed on the surface of the car, it can guide the designer to optimize the shape of the car and improve the design scheme. This dataset contains a total of 889 items of data, of which 789 items are the training set and the remaining 100 data are the test set.

The complete point cloud in each data in this dataset contains 32,186 mesh points. Of these, 3,268 points are discretized from the shape of the car, and the remaining points are in the space around the car. The former contains the wind pressure $p$ distributed on the surface of the car, and the latter contains the surrounding air flow velocity $v$. This dataset is available at: \url{http://www.nobuyuki-umetani.com/publication/mlcfd_data.zip}.

\subsection*{Heat2d}

Heat2d is a complex heat conduction dataset with multi-scale meshes, complex geometries, and dynamic boundary conditions, designed by GNOT's team \cite{hao2023gnot}. There are two versions of the dataset, the full version contains 5,500 data, 5,000 of which are the training set and the remaining 500 are the test set. A reduced version contains 1,100 data points, with 1,000 for the training set and 100 for the test set. In the original paper of GNOT, the authors conducted experiments under two versions and achieved 2.56e-2 and 4.13e-2 relative $L_2$ error, respectively. Since the authors only open sourced the version with 1,100 data, our experiments were trained directly following the original data provided by them. This dataset is available at: \url{https://drive.google.com/drive/folders/1kicZyL1t4z6a7B-6DJEOxIrX877gjBC0}.

\subsection*{Beam2d}

Beam2d \cite{bai2023physics} is a classical elastic problem to predict the displacement $u$, $v$ and stress $\sigma_x$, $\sigma_y$, $\tau_{xy}$ of a cantilever beam subjected to a bending moment at its end.

\begin{figure}[thpb]
	\centering
    \includegraphics[scale=0.5, trim=0 0 0 0, clip]{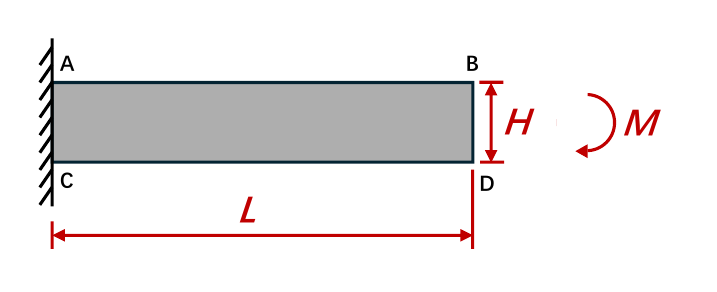}
	\caption{Elastic problems in beam bending.}
	\label{fig:beam}
\end{figure}

The analytical solution of displacement $u$ and $v$ in this problem has been given by:
\begin{equation}
    \begin{aligned}
	   u(x,y)&=\frac{3M(2\mu+\lambda)}{\mu(\mu+\lambda)LH^3}xy,\\
          v(x,y)&=-\frac{3M}{2\mu(\mu+\lambda)LH^3}((2\mu+\lambda)x^2+\lambda y^2),
    \end{aligned}
\end{equation}
where we set the length $L=5$ and height $H=1$, the Lamé constants $\mu$ and $\lambda$ can be calculated as:
\begin{equation}
    \begin{aligned}
	   \mu&=\frac{E}{2(1+\nu)},\\
          \lambda&=\frac{E\nu}{(1+\nu)(1-2\nu)},
    \end{aligned}
\end{equation}
where the Young's modulus $E$ and Poisson's ratio $\nu$ are the properties of materials, we set to $E=20GPa$ and $\nu=0.3$. The analytical solution for stress components $\sigma_x$, $\sigma_y$, and $\tau_{xy}$ is given by:
\begin{equation}
    \sigma_x = \frac{M}{I} y, \quad \sigma_y = 0, \quad \tau_{xy} = 0,
\end{equation}
where $M$ is the applied bending moment at the free end of the beam, $I$ is the second moment of inertia of the beam's cross-section, and $y$ is the distance from the neutral axis (centerline of the cross-section).

For a rectangular cross-section of the beam with width $B$ and height $H$, the second moment of inertia $I$ is:
\begin{equation}
    I = \frac{1}{12} B H^3.
\end{equation}

Substituting $I$ into the equation for $\sigma_x$, the stress components become:
\begin{equation}
    \sigma_x = \frac{12 M}{B H^3} y, \quad \sigma_y = 0, \quad \tau_{xy} = 0.
\end{equation}

When training each model, we output all five variables. However, since the ground truth of $\sigma_y$ and $\tau_{xy}$ are all equal to 0 in this case, their $L_2$ error will be very large. We thereby evaluate the joint error of all the three stresses by calculating the value of von Mises yield stress $\sigma_\text{v}$, which is calculated as:
\begin{equation}
    \sigma_{\text{v}} = \sqrt{\sigma_x^2 + \sigma_y^2 - \sigma_x \sigma_y + 3\tau_{xy}^2}.
\end{equation}

\begin{figure}[thpb]
	\centering
    \includegraphics[scale=0.5, trim=0 0 0 0, clip]{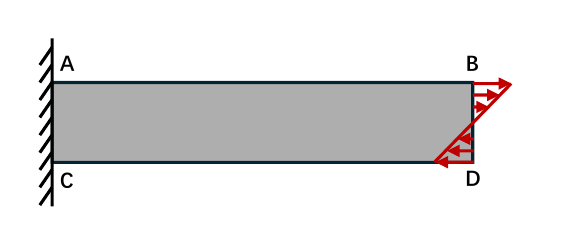}
	\caption{Equivalent load for the moment on the side of the beam.}
	\label{fig:beam1}
\end{figure}

In this experiment, we model the end bending moment as an equivalent distributed load on the boundary BD, as shown in Figure \ref{fig:beam1}. Consequently, this dataset has dynamic boundary conditions. We randomly initialized 1,000 training data, each containing 1,000 completely random sampling points and with different bending moments $M$. The test set contains 100 items of data, each containing 5,000 completely random sampling points. Therefore, the sequence lengths at training and query phases are inconsistent.

\begin{figure}[thpb]
	\centering
    \includegraphics[width=0.6\linewidth, trim=0 0 0 0, clip]{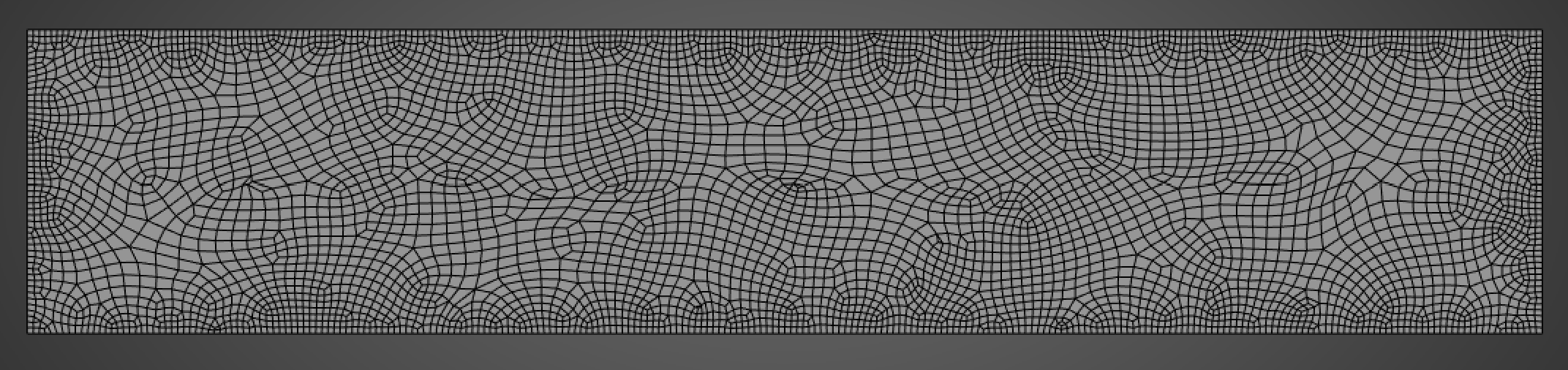}
	\caption{Mesh of the beam from Ansys.}
	\label{fig:mesh}
\end{figure}

We also provide the FEM mesh with 5,404 nodes fabricated by Ansys Workbench, as shown in Figure \ref{fig:mesh}.

\subsection*{Heatsink2d}

\begin{figure}[thpb]
	\centering
    \includegraphics[width=0.8\linewidth, trim=0 0 0 0, clip]{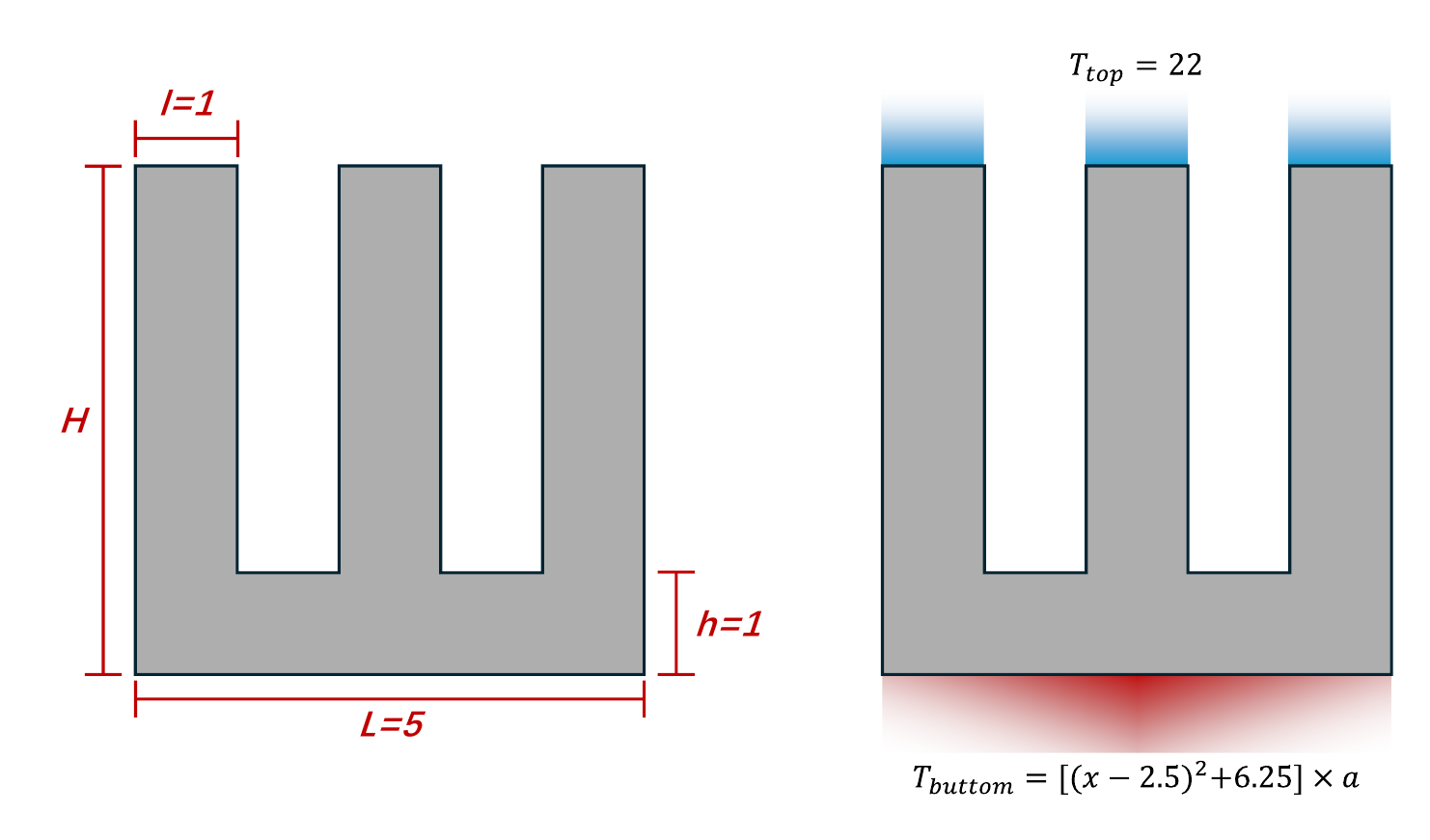}
	\caption{Definition of Heatsink2d dataset.}
	\label{fig:heatsink}
\end{figure}

Heatsink2d is a complex thermodynamic problem with multiple inputs, multiple scales, and dynamic geometry, which is designed by us. The goal is to solve for the temperature $T$ distribution inside a 2D heat sink. This problem is governed by the steady-state heat conduction equation:
\begin{equation}
    \frac{\partial^2 T}{\partial x^2} + \frac{\partial^2 T}{\partial y^2} + Q(x, y) = 0,
\end{equation}
where $T$ is the temperature distribution, and $Q$ is the heat source term per unit volume. Its geometric structure is shown in Figure \ref{fig:heatsink}. Among them, The temperature distribution at the bottom of the heat sink is given by:
\begin{equation}
    T_{\text{bottom}}=[(x-2.5)^2+6.25]\times a,
\end{equation}
where $a$ is a preset coefficient, and different instances have different heights $H$ and coefficient $a$.

The primary challenge of this dataset lies in using only the PDE and boundary conditions as the loss function to guide model convergence, without incorporating any labeled data. Previous studies \cite{unknown} have demonstrated that learning directly from PDEs is inherently more challenging and highly dependent on the model's stability. For the steady heat conduction equation $-\nabla\!\cdot\!(k\nabla T)=Q$, we use:
\begin{equation}
    r_{\text{PDE}}(x,y) \;=\; -\,\nabla\!\cdot\!\big(k\,\nabla T(x,y)\big) \;-\; Q(x,y).
\end{equation}

In our experiments, we set $k\!=\!1$; when neglecting the internal heat source, therefore take $Q\!\equiv\!0$. We define the Dirichlet and Neumann (heat–flux) boundary residuals as:
\begin{equation}
    \begin{aligned}
        r_{\text{BC},\text{bottom}}(x) &= T(x,0) - T_{\text{bottom}}(x), \\
        r_{\text{BC},\text{top}}(x)    &= T(x,H) - T_{\text{top}}, \\
        r_{\text{BC},\text{other}}(x,y) &= \nabla T(x,y)\!\cdot\!\mathbf{n}(x,y) - q_n(x,y),
    \end{aligned}
\end{equation}
where $\mathbf{n}$ is the outward unit normal to the boundary and $q_n$ is the prescribed normal heat flux (for adiabatic boundaries, $q_n \equiv 0$). Here, “other” denotes the Neumann part of the boundary.

Therefore, the loss function is given by:
\begin{small}
    \begin{equation}
        \begin{aligned}
            \mathcal{L} \;&=\; \mathcal{L}_{\text{PDE}} + \mathcal{L}_{\text{BC}}\\
            \;&=\; \sum r_{\text{PDE}}^{\,2}
                + \sum r_{\text{BC},\text{bottom}}^{\,2}
                + \sum r_{\text{BC},\text{top}}^{\,2}
                + \sum r_{\text{BC},\text{other}}^{\,2}.
        \end{aligned}
    \end{equation}
\end{small}

\begin{figure}[thpb]
	\centering
    \includegraphics[width=0.4\linewidth, trim=0 0 0 0, clip]{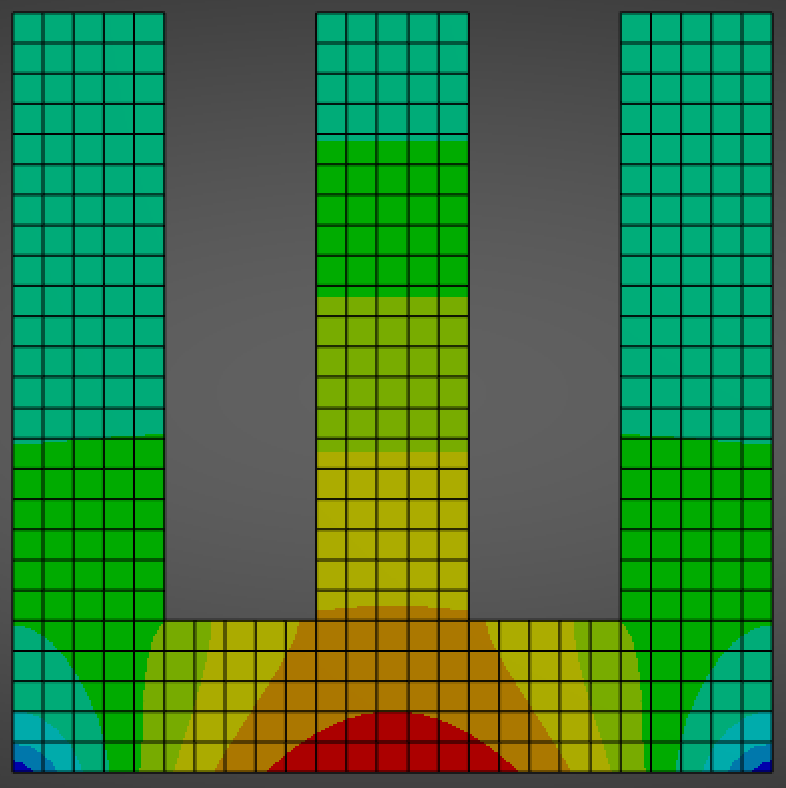}
	\caption{FEM result from Ansys.}
	\label{fig:heatsink_r}
\end{figure}

In this experiment, the heights $H$ and coefficients $a$ are randomly initialized in each epoch. The test set consists of FEM results obtained using Ansys Workbench, comprising 10 distinct datasets. The results are presented in Figure \ref{fig:heatsink_r}.

\section*{Appendix C. Details in Ablation Experiment}

\begin{figure}[thpb]
	\centering
    \includegraphics[width=0.8\linewidth, trim=0 0 0 0, clip]{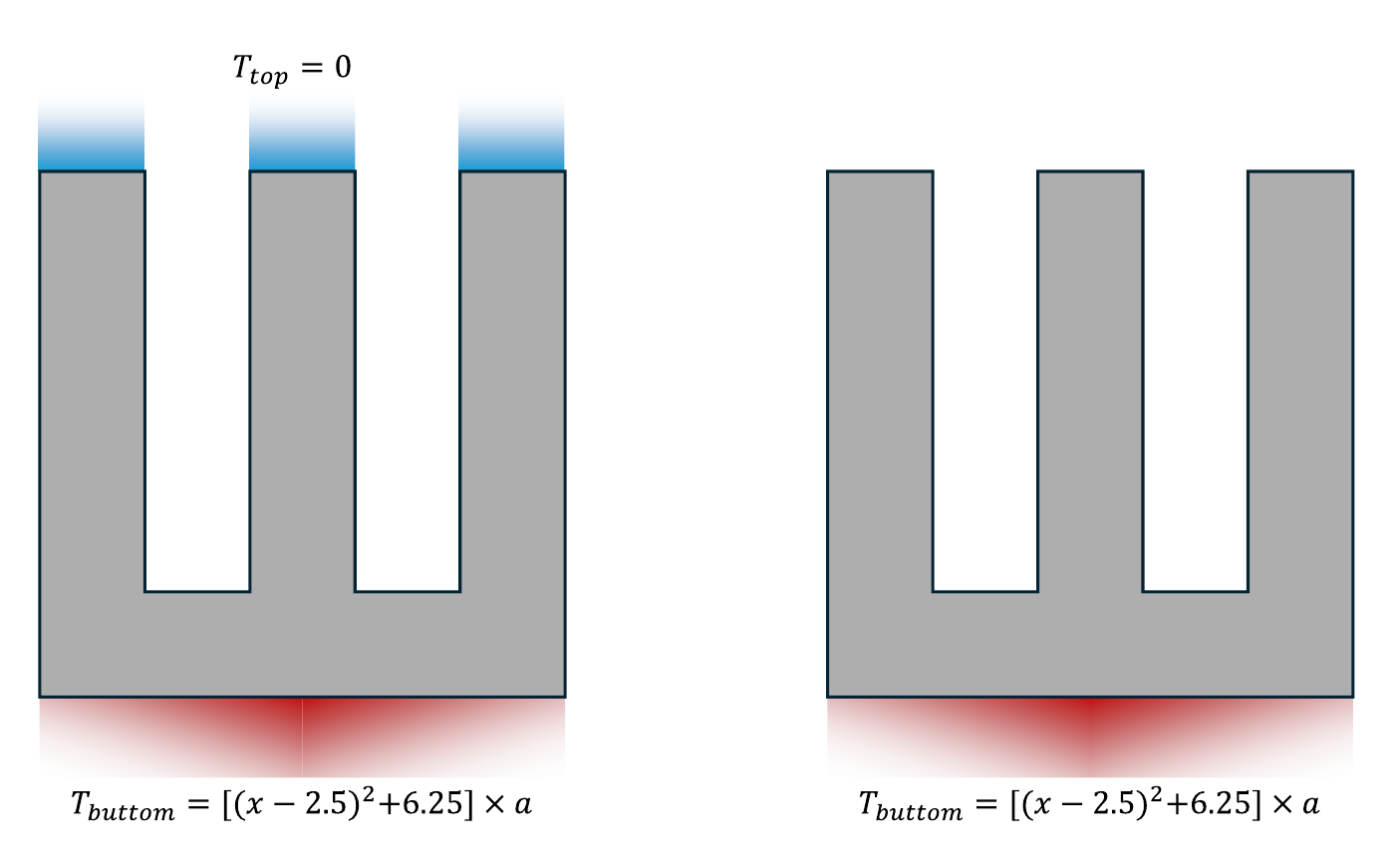}
	\caption{Two different types of boundaries.}
	\label{fig:xr3}
\end{figure}

To assess the significance of the GCE layer, we designed an experiment modified from HeatSink2d dataset, where a boundary is represented as a heat flux (Neumann) boundary in part of the dataset and as a zero-value temperature (Dirichlet) boundary in the remainder. This small dataset contains 10 data, 5 of which are Dirichlet boundaries and the other 5 are Neumann boundaries. As shown in Figure \ref{fig:xr3}.

For the different boundary conditions, we encode them into one vector, as below:
\begin{equation}
    \Vec{v}_{\text{BC}}=[T, \Vec{n}_x, \Vec{n}_y,\partial T/\partial x,\partial T/\partial y],
\end{equation}
where $T$ is the surface temperature for Dirichlet boundary, $\Vec{n}_x$ and $\Vec{n}_y$ are the surface normal vector for Neumann boundary, and $\partial T/\partial x$ and $\partial T/\partial y$ are the first derivative of Temperature for Neumann boundary. If the boundary conditions are of Dirichlet type, then the $\Vec{n}_x$, $\Vec{n}_y$, $\partial T/\partial x$ and $\partial T/\partial y$ are all masked with 0, and vice versa.

We consider the top boundary, when the boundary is Dirichlet type and temperature value is 0, the input vector is $[0, 0, 0, 0, 0]$ , and when the boundary is Neumann type, the input vector will also be $[0, 0, 0, 0, 0]$. It can be seen that for the neural network, the inputs of different boundary types under this situation are exactly the same. However, they correspond to completely different outputs.

To verify the ability of the GCE layer to handle this ambiguity, we evaluate the performance of the plain MLP layer, the MLP layer with an added boundary type indicator, and our GCE layer in these two partial datasets, respectively. Experimental results show that our GCE layer can effectively improve the accuracy of the model in the presence of input ambiguity.

\section*{Appendix D. Experimental Setup}

\subsection*{Training Hyperparameters}

The hyperparameters for each model in training phase are shown in Table \ref{tab:hp1}.

\begin{table*}
    \centering
    \begin{small}
        \begin{tabular}{l|cccccc}
            \toprule
            \textbf{Hyperparam} & Poisson & Darcy Flow & Shape-Net Car & Heat2d & Beam2d & Heatsink2d \\
            \midrule
            Optimizer       & L-BFGS & AdamW & AdamW & Adam + L-BFGS & L-BFGS & L-BFGS \\
            Learning Rate   & 5e-1 & 1e-3 & 1e-3 & 1e-3 + 3e-3 & 5e-1 & 5e-1 \\
            Epoch Number    & 50  & 200 & 200 & 100 + Until Convergence & Until Convergence & Until Convergence \\
            Batch Size      & 1 & 2 & 2 & 50 & 30 & 30 \\
            \bottomrule
        \end{tabular}
    \end{small}
    \caption{Training hyperparameters for each datasets. To ensure fairness, this configuration is adopted for all models.}
    \label{tab:hp1}
\end{table*}

\subsection*{Model Configuration}

The model configurations of the different methods are shown in Table \ref{tab:mmet}, \ref{tab:mmete}, \ref{tab:pinnsf}, \ref{tab:transo}, and \ref{tab:gnot}.

\begin{table*}
    \centering
    \begin{small}
    \begin{tabular}{l|ccccccc}
        \toprule
        \textbf{Datasets} & d\_emb & patch\_size & patch\_order & d\_model & n\_encoder & n\_decoder & n\_head \\
        \midrule
        Poisson         & 16 & 4   & 16 & 32  & 2 & 2 & 1 \\
        Darcy Flow      & 32 & 2   & 16 & 128 & 2 & 2 & 2 \\
        Shape-Net Car   & 32 & 2   & 16 & 128 & 2 & 2 & 2 \\
        Heat2d          & 32 & 1   & 16 & 192 & 2 & 2 & 3 \\
        Beam2d          & 32 & 128 & 16 & 128 & 2 & 2 & 2 \\
        Heatsink2d      & 32 & 64  & 16 & 128 & 2 & 2 & 2 \\
        \bottomrule
    \end{tabular}
    \end{small}
    \caption{Model configuration of MMET in all the datasets. Where d\_emb is the dimension of the GCE layer, patch\_size is the size of patch embedding, patch\_order is the order of Hilbert curve, d\_model is the dimension of the encoder and decoder block, n\_encoder is the number of encoder block, n\_decoder is the number of decoder block, and n\_head is the number of attention head.}
    \label{tab:mmet}
\end{table*}

\begin{table*}
    \centering
    \begin{small}
    \begin{tabular}{l|ccccccc}
        \toprule
        \textbf{Datasets} & d\_emb & patch\_size & patch\_order & d\_model & n\_encoder & n\_head \\
        \midrule
        Poisson         & 16 & 4   & 16 & 32  & 4 & 1 \\
        Darcy Flow      & 32 & 2   & 16 & 128 & 4 & 2 \\
        Shape-Net Car   & 32 & 2   & 16 & 128 & 4 & 2 \\
        Heat2d          & 32 & 1   & 16 & 192 & 4 & 3 \\
        \bottomrule
    \end{tabular}
    \end{small}
    \caption{Model configuration of MMET (encoder-only) in all the datasets. Where d\_emb is the dimension of the GCE layer, patch\_size is the size of patch embedding, patch\_order is the order of Hilbert curve, d\_model is the dimension of the encoder and decoder block, n\_encoder is the number of encoder block, and n\_head is the number of attention head.}
    \label{tab:mmete}
\end{table*}

\begin{table*}
    \centering
    \begin{small}
    \begin{tabular}{l|cccc}
        \toprule
        \textbf{Datasets} & d\_model & d\_hidden & n\_layer & n\_head \\
        \midrule
        Poisson         & 32  & 32    & 4 & 1  \\
        Darcy Flow      & 128 & 128   & 4 & 2  \\
        Shape-Net Car   & 128 & 128   & 4 & 2  \\
        \bottomrule
    \end{tabular}
    \end{small}
    \caption{Model configuration of PinnsFormer in all the datasets. Where d\_model is the dimension of the encoder and decoder block, d\_hidden is the dimension of the output MLP layer, n\_layer is the number of encoder and decoder block, and n\_head is the number of attention head.}
    \label{tab:pinnsf}
\end{table*}

\begin{table*}
    \centering
    \begin{small}
    \begin{tabular}{l|cccccc}
        \toprule
        \textbf{Datasets} & d\_model & d\_hidden & n\_layer & n\_head & slice\_num & MLP\_ratio \\
        \midrule
        Poisson         & 32  & 32    & 4 & 1 & 8  & 1 \\
        Darcy Flow      & 128 & 128   & 4 & 2 & 32 & 1 \\
        \bottomrule
    \end{tabular}
    \end{small}
    \caption{Model configuration of Transolver in all the datasets. Where d\_model is the dimension of the encoder and decoder block, d\_hidden is the dimension of the output MLP layer, n\_layer is the number of encoder and decoder block, n\_head is the number of attention head, slice\_num is the number of slice in physics attention, and MLP\_ratio is the expand ratio of the inner layer of MLPs.}
    \label{tab:transo}
\end{table*}

\begin{table*}
    \centering
    \begin{small}
    \begin{tabular}{l|cccccc}
        \toprule
        \textbf{Datasets} & trunk\_size & branch\_sizes & n\_experts & d\_model & n\_layer & n\_head \\
        \midrule
        Poisson         & 2 & -           & 0 & 32  & 4 & 1 \\
        Darcy Flow      & 2 & -           & 0 & 128 & 4 & 2 \\
        Beam2d          & 2 & [2, 4, 6]   & 3 & 128 & 4 & 2 \\
        Heatsink2d      & 2 & [2, 3, 6]   & 3 & 128 & 4 & 2 \\
        \bottomrule
    \end{tabular}
    \end{small}
    \caption{Model configuration of GNOT in all the datasets. Where trunk\_size is the input dimension of the trunk network, branch\_sizes is the input dimension of the branch network, n\_experts is the number of the expert network, d\_model is the dimension of the encoder and decoder block, n\_layer is the number of encoder and decoder block, and n\_head is the number of attention head.}
    \label{tab:gnot}
\end{table*}

\end{document}